\documentclass[tightenlines, twocolumn, notitlepage, nofootinbib, superscriptaddress]{revtex4-2}

\usepackage[dvipsnames]{xcolor}
\usepackage{amsmath}
\usepackage{amssymb}
\usepackage{bibunits}
\usepackage{changes}
\usepackage{enumerate}
\usepackage{floatrow}
\usepackage{hyperref}
\usepackage[symbol]{footmisc}
\usepackage{graphicx}
\usepackage{units}
\usepackage{xspace}
\usepackage{booktabs}
\usepackage{listings}
\usepackage[frozencache,cachedir=.]{minted}
\usepackage{booktabs}
\usepackage{multirow}
\usepackage[utf8x]{inputenc}

\usepackage{array}
\newcolumntype{P}[1]{>{\centering\arraybackslash}p{#1}}
\newcolumntype{M}[1]{>{\centering\arraybackslash}m{#1}}

\graphicspath{{./}, {figures/}}
\setcitestyle{super}

\newcommand{\olympuslabel}[1]{\texttt{#1}\xspace}

\newcommand{\coco}{\textsc{Coco}\xspace}
\newcommand{\olympus}{\textsc{Olympus}\xspace}
\newcommand{\sherpa}{\textsc{Sherpa}\xspace}
\newcommand{\optuna}{\textsc{Optuna}\xspace}
\newcommand{\pygmo}{\textsc{Pygmo}\xspace}

\newcommand{\basin}{\textsc{Basin Hopping}\xspace}
\newcommand{\cg}{\textsc{Conjugate Gradient}\xspace}
\newcommand{\cma}{\textsc{CMA-ES}\xspace}

\newcommand{\genetic}{\textsc{Genetic}\xspace}
\newcommand{\devol}{\textsc{Differential Evolution}\xspace}
\newcommand{\gpyopt}{\textsc{GPyOpt}\xspace}
\newcommand{\grid}{\textsc{Grid Search}\xspace}
\newcommand{\hyperopt}{\textsc{HyperOpt}\xspace}
\newcommand{\latin}{\textsc{Latin Hypercube}\xspace}
\newcommand{\lbfgs}{\textsc{LBFGS}\xspace}
\newcommand{\pso}{\textsc{Particle Swarms}\xspace}
\newcommand{\phoenics}{\textsc{Phoenics}\xspace}
\newcommand{\random}{\textsc{Random Search}\xspace}
\newcommand{\simplex}{\textsc{Simplex}\xspace}
\newcommand{\slsqp}{\textsc{SLSQP}\xspace}
\newcommand{\snobfit}{\textsc{Snobfit}\xspace}
\newcommand{\sobol}{\textsc{Sobol Sequence}\xspace}
\newcommand{\steep}{\textsc{Steepest Descent}\xspace}
\newcommand{\summit}{\textsc{Summit}\xspace}
\newcommand{\openaigym}{\textsc{OpenAI Gym}\xspace}

\definechangesauthor[name=Florian, color=orange]{fh}

\definechangesauthor[name=Riley, color=blue]{rh}

\definechangesauthor[name=Matteo, color=brown]{ma}

\makeatletter
\renewcommand*{\p@subsection}{\thesection.}
\makeatother

\usepackage{booktabs,arydshln}

\makeatletter
\def\adl@drawiv#1#2#3{%
        \hskip.5\tabcolsep
        \xleaders#3{#2.5\@tempdimb #1{1}#2.5\@tempdimb}%
                #2\z@ plus1fil minus1fil\relax
        \hskip.5\tabcolsep}
\newcommand{\cdashlinelr}[1]{%
  \noalign{\vskip\aboverulesep
           \global\let\@dashdrawstore\adl@draw
           \global\let\adl@draw\adl@drawiv}
  \cdashline{#1}
  \noalign{\global\let\adl@draw\@dashdrawstore
           \vskip\belowrulesep}}
\makeatother

\begin{document}

	%=== TITLEPAGE ===%

	\title{\large{Olympus: a benchmarking framework for noisy optimization and experiment planning}}

	\date{\today}

	\author{Florian H\"ase}
	\thanks{These authors contributed equally}
	\affiliation{Department of Chemistry and Chemical Biology, Harvard University, Cambridge, Massachusetts, 02138, USA}
	\affiliation{Vector Institute for Artificial Intelligence, Toronto, ON M5S 1M1, Canada}
	\affiliation{Department of Computer Science, University of Toronto, Toronto, ON M5S 3H6, Canada}
	\affiliation{Chemical Physics Theory Group, Department of Chemistry, University of Toronto, Toronto, ON M5S 3H6, Canada}
	\affiliation{Atinary Technologies S\`arl, 1006 Lausanne, VD, Switzerland}
	\author{Matteo Aldeghi}
	\thanks{These authors contributed equally}
	\affiliation{Vector Institute for Artificial Intelligence, Toronto, ON M5S 1M1, Canada}
	\affiliation{Department of Computer Science, University of Toronto, Toronto, ON M5S 3H6, Canada}
	\affiliation{Chemical Physics Theory Group, Department of Chemistry, University of Toronto, Toronto, ON M5S 3H6, Canada}
	\author{Riley J. Hickman}
	\affiliation{Department of Computer Science, University of Toronto, Toronto, ON M5S 3H6, Canada}
	\affiliation{Chemical Physics Theory Group, Department of Chemistry, University of Toronto, Toronto, ON M5S 3H6, Canada}
	\author{Lo\"ic M. Roch}
	\affiliation{Vector Institute for Artificial Intelligence, Toronto, ON M5S 1M1, Canada}
	\affiliation{Department of Computer Science, University of Toronto, Toronto, ON M5S 3H6, Canada}
	\affiliation{Chemical Physics Theory Group, Department of Chemistry, University of Toronto, Toronto, ON M5S 3H6, Canada}
	\affiliation{Atinary Technologies S\`arl, 1006 Lausanne, VD, Switzerland}
	\author{Melodie Christensen}
	\affiliation{Process Research and Development, Merck \& Co., Inc., Rahway, NJ, USA}
    \affiliation{Department of Chemistry, The University of British Columbia, Vancouver, V6T 1Z1, Canada}
    \author{Elena Liles}
    \affiliation{Department of Chemistry, The University of British Columbia, Vancouver, V6T 1Z1, Canada}
    \author{Jason E. Hein}
    \affiliation{Department of Chemistry, The University of British Columbia, Vancouver, V6T 1Z1, Canada}
	\author{Al\'an Aspuru-Guzik}
	\email{alan@aspuru.com}
	\affiliation{Vector Institute for Artificial Intelligence, Toronto, ON M5S 1M1, Canada}
	\affiliation{Department of Computer Science, University of Toronto, Toronto, ON M5S 3H6, Canada}
	\affiliation{Chemical Physics Theory Group, Department of Chemistry, University of Toronto, Toronto, ON M5S 3H6, Canada}
	\affiliation{Lebovic Fellow, Canadian Institute for Advanced Research, Toronto, Ontario M5G 1Z8, Canada}

	%=== ABSTRACT ===%
	
\begin{abstract}

Research challenges encountered across science, engineering, and economics can frequently be formulated as optimization tasks.
In chemistry and materials science, recent growth in laboratory digitization and automation has sparked interest in optimization-guided autonomous discovery and closed-loop experimentation.
Experiment planning strategies based on off-the-shelf optimization algorithms can be employed in fully autonomous research platforms to achieve desired experimentation goals with the minimum number of trials.
However, the experiment planning strategy that is most suitable to a scientific discovery task is \textit{a priori} unknown while rigorous comparisons of different strategies are highly time and resource demanding. 
As optimization algorithms are typically benchmarked on low-dimensional synthetic functions, it is unclear how their performance would translate to noisy, higher-dimensional experimental tasks encountered in chemistry and materials science.
We introduce \olympus, a software package that provides a consistent and easy-to-use framework for benchmarking optimization algorithms against realistic experiments emulated \emph{via} probabilistic deep-learning models. 
\olympus includes a collection of experimentally derived benchmark sets from chemistry and materials science and a suite of experiment planning strategies that can be easily accessed \emph{via} a user-friendly \textit{python} interface. 
Furthermore, \olympus facilitates the integration, testing, and sharing of custom algorithms and user-defined datasets.
In brief, \olympus mitigates the barriers associated with benchmarking optimization algorithms on realistic experimental scenarios, promoting data sharing and the creation of a standard framework for evaluating the performance of experiment planning strategies.

\end{abstract}

	\maketitle

	%=== INTRODUCTION ===%

\section{Introduction}

    % general paragraph on optimization
    % - define optimization
    % - motivate that different tasks pose different requirements
    % - highlight the importance of benchmarks
    Optimization tasks are ubiquitous across science, engineering, and economics.
    They typically involve the identification of specific choices for controllable parameters under which a system of interest yields a desired response. 
    The development of efficient strategies that lead to the discovery of such optimal parameter choices is of significant importance and has long been of interest to many scientific communities.
    Selecting an appropriate optimization strategy for a problem with unknown structure is non-trivial given that a single, overall superior strategy does not exist.\cite{Wolpert:1997, Droste:2002}
    Specifically, the qualities of a single optimization strategy including convergence, computational demand, or requirements on the function to be optimized, could be ideal for some applications but render the same strategy inapplicable to other tasks.
    Understanding the challenges of optimization tasks in specific domains and the behavior of different algorithms for such tasks is essential to the development of efficient search strategies that are suitable to the considered application. 
    Empirical assessments of the performance of different optimization strategies on realistic and domain-relevant scenarios is thus of paramount practical relevance.

    % optimization in the context of materials science and chemistry
    % - explain materials discovery as an optimization task
    % - highlight fundamental challenges of this optimization task (while neglecting the fact that we often need to optimize for more than one property)
    One aspect where optimization has recently gained increased attention is the digitization of scientific discovery with autonomous platforms.\cite{Wilbraham:2020, Krishnamurth:2018, Aspuru:2018}
    The emergence of ever more sophisticated and reliable automated experimentation equipment in chemistry and materials science over the last decades has increasingly allowed for formulation of scientific discovery as an optimization task.\cite{Le:2016, Houben:2015, Danielson:1997}
    In this formulation, compositions of candidate materials and processing conditions to fabricate multi-component materials are optimized to reach desired goals with respect to the physical and chemical properties of the synthesized material. 
    Key missions in these fields relate to the discovery of functional molecules and advanced materials to tackle societal challenges such as climate change, renewable energy, sustainability, or clean water, which can be directly approached by modifying the structures and compositions of candidate materials to optimize their physical and chemical properties.\cite{Tabor:2018}
    
    Automated instrumentation is now being combined with data-driven optimization strategies to enable autonomous molecule and materials development in self-driving laboratories.\cite{Hase:2019_trends} Autonomous experimentation leverages these data-driven strategies to suggest molecules or materials candidates that are synthesized and characterized by robotic platforms,\cite{Hase:2019_trends, Stein:2019, Coley:2019_autonomous, Jensen:2019_autonomous} with real-time feedback on the suggested candidates being collected in the form of physical or chemical measurements.
    In this vision, the experimentation process requires minimal human intervention once the experimental campaign has been defined.
    The integration of algorithmic experiment planners with robotic hardware into an autonomous platform has already been shown to substantially lower the development costs of organic photovoltaic materials,\cite{Langner:2020} identify novel chemical reactions,\cite{Granda:2018} yield unexpected findings for thin film technologies,\cite{Macleod:2020} the discovery of photocatalysts for hydrogen production from water,\cite{Burger:2020} and mechanical design,\cite{Gongora:2020} amongst other applications.

    % motivate optimization for materials science and chemistry and physics 
    Several different optimization strategies have already been used for automated scientific discovery.
    While some of these optimization algorithms have been designed for broad applicability across general optimization tasks, other approaches have been developed with the more specific goal of planning laboratory experiments and are based on assumptions about the expected experimental response surfaces.
    For example, Design of experiments (DoE) constitutes a frequently employed strategy to identify optimal conditions for chemical reactions,\cite{Reizman:2016, Ingham:2014} where the system of interest is probed on a grid of different parameter choices.
    Chemical reactions have also been optimized with the \snobfit algorithm,\cite{Krishnadasan:2007, Walker:2017, Bedard:2018} variants of the \simplex method,\cite{Fitzpatrick:2016, Cortes:2018, Mcmullen:2010} or even gradient-based strategies.\cite{Mcmullen:2010}
    Bayesian optimization frameworks have been demonstrated on materials science applications, most often realized using Gaussian processes\cite{Xue:2016, Noack:2019, Wigley:2016} or random forests.\cite{Nikolaev:2016} 
    
    While the experiment planning strategies deployed in the aforementioned examples enabled autonomous workflows, it is not clear whether they are the most efficient ones for the considered task. 
    In fact, it has recently been reported that ill-chosen planners can increase the budget requirements for scientific discovery in the context of materials science by up to an order of magnitude.\cite{Rohr:2020} 
    Without comprehensive benchmarks, availability and ease-of-use might be the primary considerations behind the choice of experiment planning strategy, while other factors such as the speed of convergence or the computational demand are neglected. 
    The lack of the ability to evaluate the effectiveness of different experiment planning strategies thus poses a major obstacle to the development of autonomous research platforms.
    
    To resolve this challenge, we propose to benchmark experiment planning strategies on probabilistic models. 
    These models can emulate noisy experimental responses after being trained on experimental data, as previously demonstrated in the context of multi-objective optimization with autonomous research platforms.\cite{Hase:2018_chimera} 
    In particular, we suggest to use Bayesian neural networks (BNNs) due to their robustness, scalability and non-local generalization capabilities. 
    The outcome (\emph{e.g.,} reaction yield, solubility, etc.) of a specific set of experimental parameters (\emph{e.g.,} concentration, temperature, etc.) can be emulated by drawing a predictive sample from the BNN, conditioned on these parameters. 
    This approach provides a viable avenue to benchmarking experiment planning algorithms in the presence of noise and on realistic, experimentally-derived response surfaces. 
    
    Following this idea of emulating experimental response based on real data, we introduce \olympus, a comprehensive software package that provides the possibility to probe the performance of experiment planning strategies on emulated experimental surfaces collected from experiments in chemistry and materials science. 
    \olympus implements a common interface to 18 different experiment planning strategies and thus simplifies the implementation of closed-loop autonomous workflows. 
    \olympus further provides a collection of 10 experimental datasets for which emulators have been trained to serve as a standard set of benchmarks, and a collection of 23 analytical surfaces which can be modulated by different sources of stochastic noise. 
    An automated benchmarking process that determines the most efficient planner for a given application is available. 
    As such, \olympus provides the means to run comprehensive comparisons of novel optimization algorithms and planning strategies to existing ones, allowing to identify the strengths and limitations of individual tools for various scientific discovery tasks. 
    Its capacity to construct probabilistic approximations to experimental surfaces from collected data, modeling both the expected response and the noise modulations, makes \olympus a realistic benchmark suite without the need for excessive and resource demanding experimentation.
    
    In the following, we summarize the datasets and emulators available through \olympus as well as the experiment planning strategies for which intuitive yet flexible interfaces have been implemented. 
    We further highlight the application programming interface of \olympus, demonstrate how individual planners can be accessed and comprehensive benchmarks constructed with only a few lines of code. 
    We conclude by providing a performance baseline comprised of a uniform random search and invite the community to develop and demonstrate more efficient experiment planning strategies on the \olympus benchmarks.

    %=== RELATED WORKS ===%
    
\section{Background and related works}

\begin{table*}[!hbt]
    \centering
    \caption{Feature comparison of different benchmarking tools. This work introduces \olympus, which targets benchmark applications related to autonomous scientific discovery in chemistry and materials science. $^{\dagger}$limited capability.}
    \label{tab:toolkit_comparison}
    \begin{tabular}{l P{1.7cm} P{1.7cm} P{1.6cm} P{1.8cm} P{1.3cm} P{2.0cm} p{4.4cm}}
        \cline{1-8}
        Toolkit  &  Optimizer interfaces & Synthetic benchmarks & Noisy surfaces & Emulated experiments & Visuali- zations & Community contributions & Primary purpose \\
        \cline{1-8}
        \coco\cite{Elhara:2019}   & no  & yes & yes & no & yes & no & Continuous optimization \\  
        \openaigym\cite{Brockman:2016} & no  & yes  & yes & no & yes & no & Reinforcement learning \\
        \sherpa\cite{Hertel:2020} & yes & no  & no  & no & yes & no & Hyperparameter optimization \\
        \optuna\cite{Akiba:2019}  & yes & no  & no  & no & yes & no & Hyperparameter optimization \\
        \pygmo\cite{Biscani:2020} & yes & yes & no  & no & yes$^{\dagger}$ & no & Parallel optimization \\
        \summit\cite{Felton:2020} & yes & yes & yes & yes & no & no & Experiment planning \\
        \cdashlinelr{1-8}  
        \olympus & yes & yes & yes & yes & yes & yes & Experiment planning \\ 
        \cline{1-8}
    \end{tabular}
\end{table*}

While the goal of an optimization task is usually well defined, the setting in which this task is approached might differ from application to application. 
Thus, the applicability of optimization strategies to certain tasks can be assessed based on multiple criteria, which are designed to highlight strengths and shortcomings of individual strategies on the considered application. 

% goal of the paragraph below: providing a broad list of libraries which mainly provide datasets and testing environments
% @everyone: Please check if we mention all relevant tools and packages here and feel free to add those that are missing
Several benchmarks and software packages have been introduced for different applications and at various levels of accessibility. 
Prominent examples in the field of machine learning include the MNIST\cite{Lecun:2010} and CIFAR-10\cite{cifar10} datasets for image recognition, which have been made available through a selection of libraries and interfaces. 
In the field of chemistry, comprehensive collections of datasets such as MoleculeNet\cite{Wu:2018} or the QMx series\cite{Blum:2009, Ruddigkeit:2012, Rupp:2012, Ramakrishnan:2014, Ramakrishnan:2015, Glavatskikh:2019} serve similar purposes. 
Frameworks such as GuacaMol\cite{Brown:2019} and Moses\cite{Polykovskiy:2018} offer benchmarking functionalities for \emph{de novo} molecular design. 
These examples provide datasets which aim to model realistic abstractions of the targeted applications on which optimization algorithms can be benchmarked.
Yet, the requirements of comprehensive benchmarking frameworks go beyond realistic use cases and also include: (i) intuitive interfaces to interact with these datasets, (ii) interfaces to established algorithms to benchmark, (iii) tools to store and analyze collected results, and (iv) the flexibility to allow the community to extend the framework with additional datasets and algorithms.  

Tab.~\ref{tab:toolkit_comparison} reports a set of currently available benchmarking toolkits for different applications. 
% describing coco: https://github.com/numbbo/coco
\coco is a platform for the systematic comparison of real-parameter global optimizers.\cite{Elhara:2019} 
It provides benchmark function testbeds, experimentation templates which are easy to parallelize, and tools for processing and visualizing data generated by one or several optimizers. 
\coco focuses on runtime as the central performance measure and optimization tasks on continuous domains with dimensionalities beyond those typically encountered in chemistry and materials science. 
% describing the OpenAI Gym: 
The \openaigym offers a series of environments and tasks to test reinforcement learning algorithms.\cite{Brockman:2016}
% describing sherpa: https://parameter-sherpa.readthedocs.io/en/latest/ ; https://openreview.net/forum?id=HklSUMyJcQ
\sherpa is a Python toolkit for hyperparameter tuning of machine learning models.\cite{Hertel:2020} As such, \sherpa offers the automated optimization of hyperparameters via a choice of hyperparameter optimization algorithms including Bayesian optimization, evolutionary approaches and Bandit/Early-stopping schemes. \sherpa orchestrates the entire optimization process and results can be visualized in a comprehensive dashboard. However, \sherpa does not provide synthetic or noisy benchmark cases.
% describing optuna: https://arxiv.org/pdf/1907.10902.pdf
\optuna is another toolkit that focuses on the optimization of hyperparameters for machine learning models.\cite{Akiba:2019} In contrast to \sherpa, \optuna implements a \emph{define-by-run} interface for a dynamic construction of search spaces. However, it also does not provide benchmark cases. 
\pygmo is a library for massively parallel optimization, which provides a unified interface to a number of gradient and heuristic based optimization algorithms, as well as to synthetic benchmark problems. 
\pygmo also provides algorithms and benchmarks for constrained and multi-objective optimization problems. 

The aforementioned software packages have been developed with ML applications in mind. \summit, however, provides a selection of chemically motivated virtual benchmarks and a selection of experiment planning strategies. Although the application space of \summit is heavily focused on reaction optimization, it targets a realistic modeling of its use cases \emph{via} physical and statistical models.
In contrast, \olympus is tailored to the needs of optimization in a broader range of experimental disciplines, including self-driving laboratories and autonomous experimentation workflows.
Specifically, it constitutes a framework to assess the algorithmic performance of data-driven experiment planning strategies in the context of autonomous experimentation for chemistry and materials science. 
It targets optimization tasks in chemistry and materials science, where the number of parameters to optimize is typically smaller than 10.
To serve this purpose, \olympus provides interfaces to optimization algorithms commonly used for experiment planning tasks and offers interfaces to noisy emulators of experimental optimization tasks. 
In addition, the benchmaking capabilities of \olympus are open to be extended by the community who can contribute their own datasets (see Sec.~\ref{sec:share_datasets}).

    %=== PACKAGE OVERVIEW ===%

\section{Package overview}

% This figure is in the wrong section just because otherwise it gets placed too late in the text
% Where do you want to place it?
% I wanted to place it close to the "package overview section" :) and ideally also separate it from other tables/figures to avoid the "sandwich" effect of multiple figures/tables on top of each other. A detail we can sort out at the end though...
    \begin{figure*}[!htb]
        \centering
        \includegraphics[width=\textwidth]{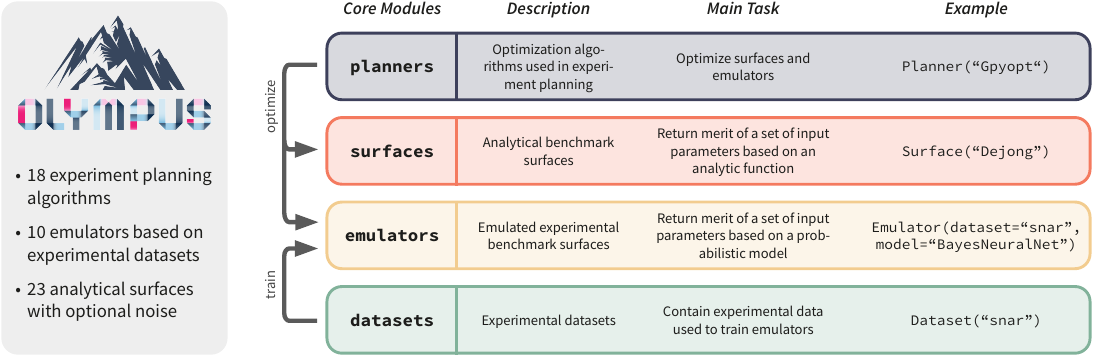}
        \caption{High-level overview of \olympus and its four core modules: (i) \texttt{planners}, which provide interfaces to common or custom optimization algorithms for experiment planning, (ii) \texttt{surfaces}, which constitute standardized interfaces to established synthetic benchmark surfaces, (iii) \texttt{emulators}, which describe a set of ML models trained to reproduce experimentally derived response surfaces encountered in chemistry and materials science, and (iv) \texttt{datasets}, which form a collection of experimental campaigns. All four core modules offer the possibility to implement and add custom methods and data. }
        \label{fig:overview_fig}
    \end{figure*}

    % brief summary of the core features of olympus
    \olympus is a modular software package that allows user interactions at different levels and can be used for data-driven experimentation as well as benchmarking experiment planning strategies. 
    With this modularity, \olympus allows for both beginner and expert use and enables performing several optimization and benchmarking tasks in a few lines of code. 
    Some common use-case scenarios are detailed in Sec.~\ref{sec:usage}, including (i) the use of different experiment planners for an autonomous workflow, (ii) benchmarking an experiment planner on an emulator, and (iii) constructing an emulator from a user-provided dataset. 
    At the heart of \olympus are four modules, \texttt{planners}, \texttt{surfaces}, \texttt{datasets}, and \texttt{emulators}, which are highlighted conceptually in this section and in Fig. \ref{fig:overview_fig}. 
    
    The \texttt{planners} module (see Sec.~\ref{sec:planners}) provides a consistent interface to 18 different experiment planning strategies via its core \texttt{Planner} class. \olympus translates a standardized access protocol to the interfaces of individual planners, making it easy to switch the experiment planning strategy of an autonomous workflow. 
    This module also provides the basis for integrating customized algorithms into the package. Available planners are listed in Table~\ref{tab:planner_list}. 
    
    The \texttt{surfaces} module provides a set of synthetic response surfaces, which are functions commonly used to evaluate and compare the performance of optimization algorithms. Similar to the \texttt{planners} module, a convenient \texttt{Surface} class allows to easily retrieve the desired analytical surface. 
    Available surfaces are listed in Table~\ref{tab:surfaces_list}. 
    While these surfaces return deterministic function evaluations by default, it is possible to pass a noise object that results in stochastic evaluations, as shown in Sec.~\ref{sec:usage_surfaces}.
    
    The \texttt{datasets} module in \olympus offers 10 core experimentally derived datasets from chemistry and materials science. These datasets vary in size and represent optimization tasks with dimensionalities from 3 to 6. The core class of this module is \texttt{Dataset}, which allows for retrieval and manipulation of the desired dataset. Available datasets are listed in Table~\ref{tab:dataset_list}. 
    Users can also load their own dataset, which can then be used to benchmark experiment planning strategies for the specific problem of interest. 
    Furthermore, users can share their datasets with the community by uploading it to the \olympus repository of user-provided datasets at \href{https://github.com/aspuru-guzik-group/olympus\_datasets}{https://github.com/aspuru-guzik-group/olympus\_datasets}. 
    Any user can then download these additional datasets to be used in \olympus \emph{via} the same interface used for the core datasets.
    
    The \texttt{emulators} module provides access to probabilistic models trained on the core \olympus datasets, reproducing the experimental responses of the corresponding experiments. With its \texttt{Emulator} class, this module also offers a high-level interface for the training of such probabilistic models on user-provided datasets. In this spirit, emulators constitute stochastic response surfaces resembling those encountered in real-life applications, thus allowing to benchmark experiment planning strategies on close-to-reality optimization tasks.
    
    %%% OVERVIEW PLANNERS %%%
    
    \subsection{Summary of included planners}
    \label{sec:planners}
        
        This section details the types of experiment planning strategies and algorithms available in \olympus and listed in Table~\ref{tab:planner_list}. More information about each specific planner can be found on the online documentation.
        
        \begin{table*}[htb]
          \begin{center} 
          \caption{List of algorithms available in \olympus. Convergence is categorized into \textit{global} (converges to global optimum), \textit{global*} (does not necessarily converge to global optimum but can overcome local minima), \textit{local} (does not necessarily converge to global optimum and does not overcome local minima).}
          \label{tab:planner_list}
            \begin{tabular}{llll}
              Planner  & Strategy & Convergence & Derivative-Free \\
            \cline{1-4}
              \gpyopt\cite{Gpyopt:2016}   & Bayesian     & global & yes \\
              \hyperopt\cite{Bergstra:2013, Bergstra:2012, Bergstra:2011} & Bayesian     & global  & yes \\
              \phoenics\cite{Hase:2018} & Bayesian     & global  & yes \\
            \cdashlinelr{1-4} 
              \genetic\cite{Fortin:2012}     & Evolutionary & global*  & yes \\
              \cma\cite{Hansen:2003, Hansen:2001}      & Evolutionary & global*  & yes \\
              \pso\cite{Eberhart:1995, Shi:1998}      & Evolutionary & global*  & yes \\
              \devol\cite{Storn:1997}     & Evolutionary & global*  & yes \\   
            \cdashlinelr{1-4} 
              \steep\cite{Curry:1944, Bouwmeester:2015}    & Gradient-based & local  & no \\
              \cg\cite{Hestenes:1952, Bouwmeester:2015}       & Gradient-based & local  & no \\
              \lbfgs\cite{Zhu:1997,Byrd:1995,Nocedal:2006}    & Gradient-based & local  & no \\
              \slsqp\cite{Kraft:1988}    & Gradient-based & local  & no \\
            \cdashlinelr{1-4} 
              \grid\cite{Anderson:2016, Box:2005, Fisher:1937}     & Grid-like  & global  & yes \\
              \latin\cite{Anderson:2016, Box:2005, Fisher:1937}    & Grid-like  & global  & yes \\
              \sobol\cite{Sobol:1967}    & Grid-like  & global  & yes \\
              \random   & Grid-like & global  & yes \\
            \cdashlinelr{1-4} 
              \snobfit\cite{Huyer:2008}  & Heuristic & global & yes \\
              \basin\cite{Wales:1997}    & Heuristic & global  & yes \\
              \simplex\cite{Nelder:1965}  & Heuristic & local  & yes \\ 
             \cline{1-4}
            \end{tabular}
            \end{center}
        \end{table*}
        
        Gradient approaches use derivative information (gradient or Hessian) at the current point to determine the location of the next point to be evaluated. Such strategies are efficient on convex optimization problems, but are not guaranteed to find the global optimum on non-convex surfaces.\cite{Bubeck:2015, Boyd:2004} 
        Most gradient-based approaches condition both the stepping direction and the step size on the local gradient. 
        The numerical approximation of gradients generally poses a challenge in the context of experimentation where the response surface is subject to noise. Nevertheless, gradient-based search strategies have been reported for the optimization of some chemical processes.\cite{Lucia:1990} 
        
        Grid-like searches constitute a more common approach to experiment planning.\cite{Anderson:2016, Box:2005, Fisher:1937} These strategies define a set of selected parameter points in the parameter space to be evaluated at the start of the optimization campaign. At every step of the campaign, the next point to be evaluated is chosen deterministically. Although grid-like searches mitigate the locality issue of gradient approaches and can reliably identify global optima, their cost scales exponentially with the dimensionality of the parameter space. Alternatives to standard full grid approaches involve the use of low discrepancy sequences, such as \latin or \sobol, to sample more effectively high dimensional spaces. The discrepancy of a sequence is considered to be low if the proportion of points falling into an arbitrary subset of the considered parameter domain is roughly proportional to the measure of this subset. Low discrepancy sequences are also known as quasi-random sequences and are commonly used to finding characteristic functions of probability density functions, higher-order moments of statistical distributions, and integration and sampling of high-dimensional deterministic functions. \random reduces the correlation between consecutive proposals even further and has been shown to be particularly effective in higher-dimensional search spaces.\cite{Bergstra:2012, Baba:1981, Matyas:1965}

        Evolutionary algorithms are population and heuristic-based approaches inspired by biological evolution.\cite{Rechenberg:1978, Schwefel:1977, Zames:1981, Koza:1992, Srinivas:1994} Each individual in the population represent a point in the search space, and their fitness corresponds to the objective evaluated at that point. Evolutionary strategies, like \cma, \pso, and \devol, evolve a population of candidate solutions simultaneously and generate new candidates based on some heuristics. The population is frequently updated, with better candidates replacing worse performing candidates.\cite{Wierstra:2014} \genetic algorithms constitute a subclass of evolutionary strategies which mimic mechanisms such as reproduction, mutation, recombination, and selection to iteratively improve the fitness of a population. 
        
        Other heuristic-based approaches are not inspired by biological evolution specifically. For example, \basin is a two-step approach that uses both local and global searches and is inspired by the energy landscape of atom clusters.\cite{Wales:1997} 
        \snobfit too combines both local and global approaches and the strategy was designed with the goal of addressing a number of practical challenges.\cite{Huyer:2008} 
        Finally, the \simplex algorithm by Nelder and Mead exploits the geometry of simplices to define an update rule that proposes new points in a downhill direction.\cite{Nelder:1965}
        
        Bayesian optimization methods are sequential, model-based approaches for the global optimization of black-box functions.\cite{Mockus:2012, Mockus:1978, Mockus:1975} 
        The function to be optimized is approximated by a surrogate model that is refined as more data is collected. 
        Based on this model, an acquisition function that evaluates the utility of candidate points can be defined, leading to the balanced exploration and exploitation of the search space of interest. Similar to evolutionary strategies, no gradient information is required and they can be used for the global optimization of black-box functions. %
        What distinguishes Bayesian optimization approaches are primarily the surrogate model and acquisition functions used. \gpyopt uses a Gaussian process to model the objective function,\cite{Gpyopt:2016} \phoenics adopts a mixture of Gaussian kernels,\cite{Hase:2018} and \hyperopt uses a tree-structured Parzen estimator.\cite{Bergstra:2011,Bergstra:2012,Bergstra:2013} 
        
       We note that the algorithms available in \olympus present different computational scalings with respect to the number of samples collected and the dimensionality of the optimization domain. These algorithmic and implementation aspects result in different runtimes and memory requirements, which  ultimately affect the applicability of each algorithm to different problems. However, given the typical problem dimensionalities ($< 10$ parameters) and runtimes ($> 10$ min per experiment) encountered experimentally, the computational cost of any of the algorithms above described is not expected to be of significant importance in autonomous workflows.

    %%% OVERVIEW DATASETS %%%

\subsection{Summary of included datasets}
\label{sec:datasets}

\begin{table*}[!htb]
    \begin{center}  
        \caption{List of the core \olympus datasets. Datasets contributed in this work are marked with $^*$.}
        \label{tab:dataset_list}
        \begin{tabular}{l p{6cm} l r r}
            Olympus label   & Topic                 & Discipline                  & \# data points & \# parameters \\
        \cline{1-5}
            \olympuslabel{alkox}$^*$    & Alkoxylation reaction                              & Organic chemistry          &   208  & 4                      \\
            \olympuslabel{colors\_bob}\cite{Roch:2020}  & Colorant mixture with 3D printed robot             & Colorimetry                &   241  & 5                \\ 
            \olympuslabel{colors\_n9}\cite{Roch:2020}   & Colorant mixture with commercial robot             & Colorimetry                &   102  & 3               \\
            \olympuslabel{fullerenes}\cite{Walker:2017:fullerenes}   & Synthesis of o-xylenyl C\textsubscript{60} adducts & Organic chemistry          &   246  & 3  \\
            \olympuslabel{hplc}\cite{Roch:2020}         & Calibration of an automated HPLC                   & Analytical chemistry       & 1,386  & 6             \\
            \olympuslabel{benzylation}\cite{Schweidtmann:2018}  & N-benzylation reaction                             & Organic chemistry          &    73  & 4       \\
            \olympuslabel{photo\_pce10}\cite{Langner:2020} & Photostability of organic photovoltaics with PCE10 polymers & Materials science & 1,040  & 4          \\
            \olympuslabel{photo\_wf3}\cite{Langner:2020}   & Photostability of organic photovoltaics with WF3 polymers & Materials science   & 1,040  & 4           \\
            \olympuslabel{snar}\cite{Schweidtmann:2018}         & Nucleophilic aromatic substitution                 & Organic chemistry          &    66  & 4      \\
            \olympuslabel{suzuki}$^*$       & Carbon-carbon cross-coupling reaction              & Organic chemistry          &   247  & 4                       \\
        \cline{1-5}
        \end{tabular}
    \end{center} 
\end{table*}

    \olympus ships with a total of ten core datasets collected from experiments spanning chemistry and materials science. The datasets have either been collected from the literature or were generated in-house. 
    With these datasets, \olympus can construct experiment emulators using probabilistic machine learning models, notably BNNs, to emulate the overall response surface of the considered experiment for an arbitrary choice of parameter values (see Sec.~\ref{sec:emulators} for details). 
    As such, the provided datasets constitute the basis for realistic benchmarks of experiment planning strategies.
    
    Table~\ref{tab:dataset_list} summarizes core information about each dataset and further details are provided in the supplementary information (see Sec.~\ref{supp_sec:datasets}). 
    All datasets are collected from experimental campaigns with three to six independently controllable parameters, one property of interest, and contain from a few tens to more than 1,000 data samples. 
    Five datasets are related to the optimization of organic chemistry reactions, one is derived from the calibration of analytical chemistry instrumentation, two address the identification of polymer blends of photovoltaic materials with favorable photodegradation properties, and two are related to the identification of the colorant mixture displaying a chosen target color. 
    This core set of datasets can be extended by community datasets contributed from individual research groups. Details are provided in Sec.~\ref{sec:share_datasets}.

    %\input{table_surfaces}

    %%% OVERVIEW EMULATORS %%%
    
\subsection{Summary of included emulators}
\label{sec:emulators}

\begin{figure*}[!htb]
    \centering
    \includegraphics[width=\textwidth]{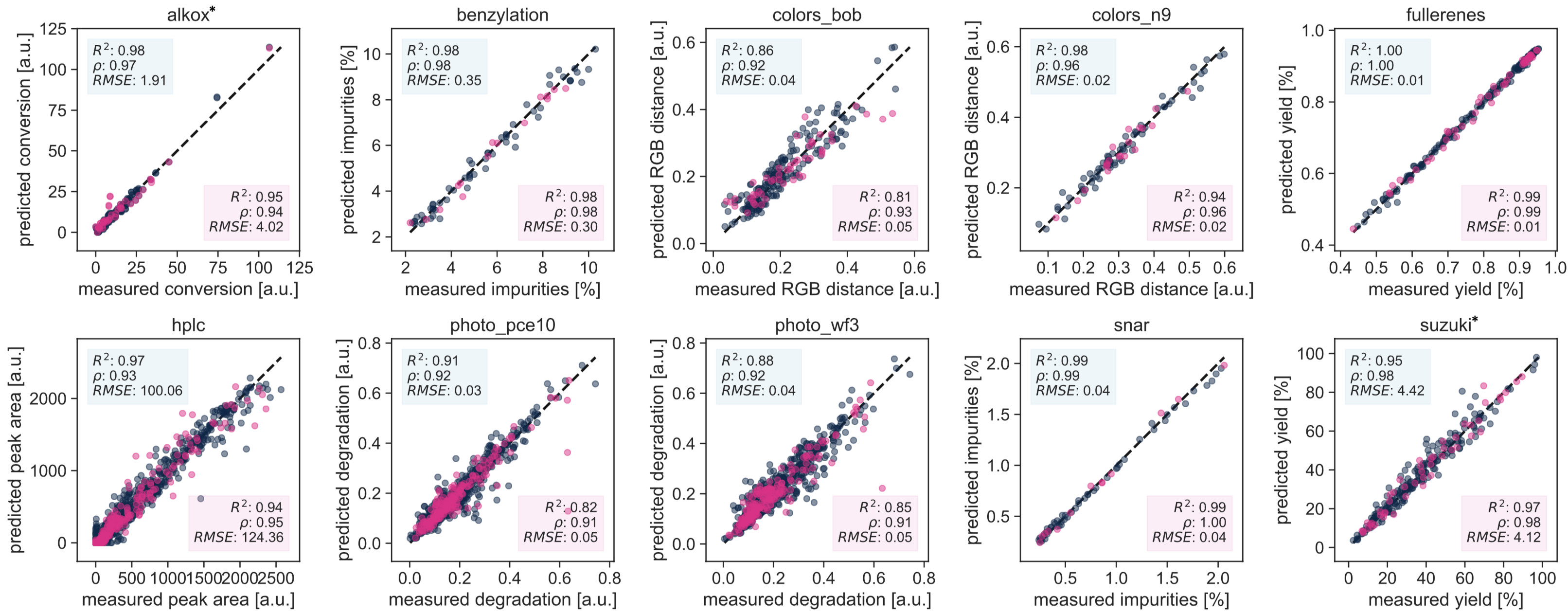}
    \caption{Parity plots with experimental versus predicted target values for all emulators based on Bayesian Neural Network models. Performance on the training set (\unit[80]\% of data; blue markers) is shown in the top-left corner of each plot and on blue background; performance on the test set (\unit[20]\% of data; pink markers) is shown in the bottom-right corner of each plot and on pink background. $R^2$ is the coefficient of determination, $\rho$ is the Spearman's rank correlation coefficient, and \textit{RMSE} is the root-mean-square error. Emulators trained on datasets introduced in this work are marked with *.}
    \label{fig:bnn_emulators}
\end{figure*}
    
The experiment emulators offered through \olympus provide a core functionality to benchmarking data-driven experiment planning strategies: the opportunity to query the response of a quasi-experimental surfaces inexpensively within milliseconds. Balancing robustness and prediction accuracy on the data-scarce datasets reported in Sec.~\ref{sec:datasets}, we construct \olympus emulators from feedforward fully connected Bayesian neural networks (BNNs). BNNs constitute probabilistic machine learning models which, contrary to standard neural network, define a distribution of possible target values conditioned on the input features. To this end, the conventional weight and bias parameters of standard neural networks are modeled as distributions themselves and the BNN is trained \emph{via} Bayesian inference. While, in theory, weights and biases can be modeled by any valid distribution, in practice the distributions are often explicitly modeled {\emph{via} a set of parameters, such as the location and scale of a normal distribution}. This approximation can greatly accelerate inference computations and make the training of a BNN overall computationally tractable. In addition to probabilistic, BNN-based emulators, we also provide deterministic, NN-based ones. 
These emulators return the same target value given a set of input features. 

Emulators are trained on \unit[80]{\%} of the data and tested on \unit[20]{\%} using a random split. The training set is furthermore split into training and validation sets for cross-validation. By default, 5 folds are used, but users can choose how many folds to use when creating their own emulators. The test set is used to probe the generalizability of the emulator. 
Model performances on both training and test sets are shown in Fig. \ref{fig:bnn_emulators} for BNN-based emulators, and in Fig. \ref{si_fig:nn_emulators} for NN-based emulators.
Emulators are constructed with different choices for hyperparameters, including the number of layers, the number of neurons per layer, activation functions, and others, which can be accessed directly through the \texttt{emulator} objects.
Activation functions for the output layer have been chosen to satisfy physical constraints, such as positivity of the property of interest. Other hyperparameters have been manually selected to achieve promising prediction accuracies. We define emulators as accurate if they achieve a Spearman's rank correlation coefficient above $0.90$ for both training and test sets, given that a monotonic relationship between predicted and measured values preserves the relative ranking of all extrema. Typical evaluations of trained emulators take less than 1~ms on a standard laptop. This cheap experiment emulation approach enable the large-scale querying of experimental responses and the rigorous benchmarking of data-driven experiment planning strategies. \\

    %%% OVERVIEW SURFACES %%%
    \subsection{Summary of included analytical surfaces}

    \begin{table*}[hbt]
        \begin{center}  
        \caption{Available analytical surfaces in \olympus.}
        \label{tab:surfaces_list}
        \begin{tabular}{l l l}
            Type                &  Property           &   Surfaces              \\
        \cline{1-3}
            Continuous          & Convex              & \texttt{Dejong}, \texttt{HyperEllipsoid}, \texttt{Zakharov} \\
                                & Non-convex          & \texttt{AckleyPath}, \texttt{Branin}, \texttt{Levy}, \texttt{Michalewicz}, \texttt{Rastrigin}, \texttt{Rosenbrock}, \texttt{Schwefel},   \\
            & & \texttt{StyblinskiTang} \\
        \cdashlinelr{1-3}
            Piece-wise constant & Convex              & \texttt{LinearFunnel}, \texttt{NarrowFunnel} \\
                                & Non-convex          & \texttt{DiscreteAckley}, \texttt{DiscreteDoubleWell}, \texttt{DiscreteMichalewicz} \\
        \cdashlinelr{1-3}
            Mixture model       & Non-convex          & \texttt{GaussianMixture}, \texttt{Denali}, \texttt{Everest}, \texttt{K2}, \texttt{Kilimanjaro}, \texttt{Matterhorn}, \texttt{MontBlanc} \\
            
        \cline{1-3}
        \end{tabular}
        \end{center} 
    \end{table*}

    In addition to experimentally-derived benchmarks, \olympus provides a suite of analytical functions traditionally used to evaluate optimization algorithms (Table \ref{tab:surfaces_list}). These functions include 11 analytic and smooth functions, such as the Branin and Rosenbrock functions, 5 piece-wise constant functions, and 6 Gaussian mixture model functions derived from a parent Gaussian mixture generator. This generator takes a number of dimensions as argument, and draws random means and covariances. By default, a full covariance matrix is drawn, but a diagonal matrix can also be requested. By fixing the random seed, the Gaussian mixture generator creates reproducible surfaces. In fact, the six Gaussian mixture models available have been obtained by fixing the random seed of each mountain-named surface (e.g. Everest) to the height of the respective mountain peak in meters (e.g. 8848). 
    
    For all analytical surfaces in \olympus, it is possible to specify noise to be added to the evaluations. In such a way, the output of these toy surfaces will be stochastic. A few commonly used noise functions, like Gaussian, uniform, and gamma-distributed noise, are already implemented and readily available in \olympus. However, custom types of noise can also be defined by the users and provided to the surface of interest, which will then return noisy evaluations. Note, that noise modulations are currently only supported for experimental responses. However, realistic experiments might also be subject to noise on the preparation of experimental parameters, such as the dispensing of desired amounts of chemicals or controlling the temperature in an experimental setup. In \olympus, users may add noise to the experimental parameters by taking advantage of the \texttt{Planners} interface.
    
    Tab.~\ref{tab:surfaces_list} summarizes the synthetic benchmark functions available in \olympus. Further details as well as illustrations of the surfaces are reported in the supplementary information (see Sec.~\ref{supp_sec:surfaces}).

    %=== HIGHLIGHT APPLICABILTY ===%
    \section{Using \olympus} 
\label{sec:usage}

    In this section, we detail the usage of \olympus on selected applications and use cases. A detailed documentation of the package is provided on GitHub.\cite{github:olympus}

    \subsection{Installation and dependencies}

    \olympus is available for download on GitHub\cite{github:olympus} and can be installed via \texttt{pip} and \texttt{conda}.
    
    \begin{minted}[mathescape,autogobble,numbersep=5pt,frame=lines,framesep=2mm, fontsize=\small]{bash}
    # Option 1 (recommended): installation via pip
    >> pip install olymp
    # Option 2: installation via anaconda
    >> conda install -c conda-forge olymp
    # Option 3: installation from source
    >> git clone https://github.com/aspuru-guzik-group/
                                            olympus.git
    >> cd olympus
    >> python setup.py develop
    \end{minted}

    The installation requires Python 3.6+ with support for \texttt{numpy} and \texttt{pandas}. However, to access specific features of the package, such as running an emulator, using specific experiment planners, or plotting the results of completed campaigns, the installation of additional packages might be required. Details are provided in the documentation.\cite{github:olympus}
    
    % USE SURFACES
    
\subsection{Evaluate analytical surfaces}
\label{sec:usage_surfaces}

The analytical surfaces in \olympus can be accessed via the \texttt{olympus.surfaces} module or the \texttt{Surface} function, with the latter loading a surface with default argument. 

\begin{minted}[mathescape,autogobble,numbersep=5pt,
               frame=lines,framesep=2mm, fontsize=\small]{python}
from olympus.surfaces import Michalewicz
surface = Michalewicz(param_dim=2, m=12)
# or, to load with default arguments
from olympus import Surface
surface = Surface("Michalewicz")
\end{minted}

The above example defines a surface with deterministic output. 
However, noise can be added to have a surface instance that returns stochastic evaluations.

\begin{minted}[mathescape,autogobble,numbersep=5pt,
               frame=lines,framesep=2mm, fontsize=\small]{python}
from olympus.surfaces import Dejong
from olympus.noises import GaussianNoise
noise = GaussianNoise(scale=0.5)
surface = Dejong(param_dim=2, noise=noise)
\end{minted}

Surfaces can then be evaluated sequentially or in batches as follows.

\begin{minted}[mathescape,autogobble,numbersep=5pt,
               frame=lines,framesep=2mm, fontsize=\small]{python}
# evaluate a single point in 2 dimensions
surface.run([0.5, 0.5])
>>> [[0.0]]
# evaluate a batch of 2 points in 2 dimensions
surface.run([[0.5, 0.5], [0.75, 0.75]])
>>> [[0.0], [3.16]]
\end{minted}

    % RUN CAMPAIGN
    
\subsection{Run a simulated campaign}
The datasets available in \olympus can be accessed via the \texttt{Dataset} class, using the keyword associated with each dataset. 

\begin{minted}[mathescape,autogobble,numbersep=5pt,
               frame=lines,framesep=2mm, fontsize=\small]{python}
# load an Olympus dataset
from olympus import Dataset
dataset = Dataset("snar")
\end{minted}

Neural Network (NN) or Bayesian Neural Network (BNN) based emulators are already available in \olympus for all datasets provided. However, the user also has the freedom to train new emulators by customising the models provided in the \texttt{olympus.models} module.

\begin{minted}[mathescape,autogobble,numbersep=5pt,
               frame=lines,framesep=2mm, fontsize=\small]{python}
from olympus import Emulator 
emulator = Emulator(dataset="snar",
                    model="BayesNeuralNet")
# or customize the model
from olympus.models import BayesNeuralNet
model = BayesNeuralNet(hidden_depth=4, 
                       out_act="sigmoid")
emulator = Emulator(dataset="snar", model=model)
\end{minted}

All algorithms described in the previous section can easily be accessed from the \texttt{olympus.planners} module or via the \texttt{Planner} function. While the former allows the user to choose specific settings for each planner, the latter loads them with default arguments.

\begin{minted}[mathescape,autogobble,numbersep=5pt,
               frame=lines,framesep=2mm, fontsize=\small]{python}
from olympus.planners import Gpyopt
planner = Gpyopt(goal="minimize",
                 model_type="GP_MCMC", 
                 acquisition_type="EI_MCMC")
# or, to load with default arguments:
from olympus import Planner
planner = Planner("Gpyopt", goal="minimize")
\end{minted}

Once a planning algorithm and an emulator have been defined, it is possible to start a simulated optimization campaign using the \texttt{optimize} method.

\begin{minted}[mathescape,autogobble,numbersep=5pt,
               frame=lines,framesep=2mm, fontsize=\small]{python}
emulator = Emulator(dataset="snar",
                    model="BayesNeuralNet")
planner = Planner("Phoenics", goal="minimize")
campaign = planner.optimize(emulator=emulator, 
                            num_iter=50)  
\end{minted}

    % TRAIN YOUR OWN EMULATOR
    \subsection{Train custom emulator}

With \olympus you can create an \texttt{Emulator} in order to generate a custom emulated response surface for a new dataset. 
For instance, if you have data for a chemical reaction of interest, for which you would like to optimize the yield, you can load the dataset from a table as follows.

\begin{minted}[mathescape,autogobble,numbersep=5pt,
               frame=lines,framesep=2mm, fontsize=\small]{python}
# load a custom dataset
from olympus import Dataset
import pandas as pd
mydata = pd.from_csv("mydata.csv")
dataset = Dataset(data=mydata, 
                  target_ids=['yield'])
\end{minted}

After this step, you can load one of the available models from the \texttt{olympus.models} module and pass it to a new \texttt{Emulator} instance, which will allow you to cross-validate and train the emulator. 
Users can override default model hyperparameters by passing custom values as arguments to the \texttt{olympus.models} module. Once you obtain an emulator with satisfactory performance, you can save it.

\begin{minted}[mathescape,autogobble,numbersep=5pt,
               frame=lines,framesep=2mm, fontsize=\small]{python}
from olympus.models import BayesNeuralNet
mymodel = BayesNeuralNet(hidden_depth=4, 
                         out_act="sigmoid")
emulator = Emulator(dataset=mydataset, 
                    model=mymodel)
emulator.cross_validate()
>>> ...

emulator.train()
>>> ...

emulator.save("my_new_emulator")
\end{minted}

    % TEST YOUR PLANNING ALGORITHM
    
\subsection{Test your planning algorithm}
\olympus allows you to create a \texttt{Planner} implementing your own custom algorithm, such that you can test it against established methods on a set of challenging benchmarks. To create such \texttt{Planner} you just need to inherit from the \texttt{CustomPlanner} class and implement the \texttt{\_ask}  method. This method should return the next query point based on the algorithm's strategy. For instance, a random sampler can be implemented as follows.

\begin{minted}[mathescape,autogobble,numbersep=5pt,
               frame=lines,framesep=2mm,fontsize=\small]{python}
from olympus.planners import CustomPlanner
from olympus import ParameterVector as PV
import numpy as np
class RandomSampler(CustomPlanner):
    def _ask(self):
        new_params = []
        for param in self._param_space:
            new_param = np.random.uniform(
                low=param['domain'][0], 
                high=param['domain'][1])
            new_params.append(new_param)
        return PV(array=new_params, 
            param_space=self.param_space)
\end{minted}

The \texttt{\_ask} method takes advantage of the \texttt{\_param\_space} attribute present in \texttt{CustomPlanner}, which is a list of dictionaries defining the parameter space over which to optimize. In addition, \texttt{\_params} and \texttt{\_values} contain the parameters and associated merits for all previous observations, respectively. These attributes will be needed for any algorithm in which the set of parameters proposed depend on the previous observations. Finally, note \texttt{\_ask} returns a \texttt{ParameterVector} object, which can be instantiated with an array or dictionary of parameters.

In the above example, an \texttt{\_\_init\_\_} method is not specified. This is because the following default one is inherited from \texttt{CustomPlanner}.

\begin{minted}[mathescape,autogobble,numbersep=5pt,
               frame=lines,framesep=2mm, fontsize=\small]{python}
def __init__(self, goal='minimize'):
    AbstractPlanner.__init__(**locals())
\end{minted}

If you would like to initialize your own \texttt{Planner} with more options, you can expand upon the above \texttt{\_\_init\_\_} method. Note it is required to keep the argument \texttt{goal} and to initialise \texttt{AbstractPlanner} as above. A tutorial on the creation of custom \texttt{Planner} classes with further details on possible customization is available as part of the online documentation.

    % DOWNLOAD/UPLOAD COMMUNITY DATASETS
    
\subsection{Download/upload community datasets}
\label{sec:share_datasets}
In addition to the set of core datasets distributed with \olympus, we allow users to share their own datasets with the community. These additional datasets are stored on GitHub and provide an extended set of benchmarks built by the autonomous experimentation community. \olympus provides intuitive command line tools to upload and download these datasets. For instance, to download the \textit{excitonics} dataset and make it available to your local \olympus installation:

\begin{minted}[mathescape,autogobble,numbersep=5pt,
               frame=lines,framesep=2mm, fontsize=\small]{text}
>> olympus download -n excitonics
\end{minted}

After the download, the \textit{excitonics} dataset will be available and you will be able to load it in the same way as the core datasets.

\begin{minted}[mathescape,autogobble,numbersep=5pt,
               frame=lines,framesep=2mm, fontsize=\small]{python}
from olympus import Dataset
dataset = Dataset("excitonics")
\end{minted}

Note that, for community-provided datasets, trained \texttt{Emulator} instances are not readily available, such that you will need to train the relevant \texttt{Emulator}.

\begin{minted}[mathescape,autogobble,numbersep=5pt,
               frame=lines,framesep=2mm,fontsize=\small]{python}
from olympus import Dataset
from olympus.models import NeuralNet
dataset = Dataset("excitonics")
model = NeuralNet(hidden_depth=3, out_act="relu")
emulator = Emulator(dataset=dataset, model=model)

emulator.train()
>>> ...

emulator.save("excitonics_nn_emulator")
\end{minted}

If you have a dataset that you think would be a useful benchmark for the community, you can upload it to the \olympus pool of datasets using the \olympus command line tools as follows.

\begin{minted}[mathescape,autogobble,numbersep=5pt,
               frame=lines,framesep=2mm,fontsize=\small]{text}
>> olympus upload -n <dataset_name> -p <dataset_path>
\end{minted}

    % PLOT RESULTS
    
\subsection{Plotting benchmark results}

\begin{figure}[!ht]
        \centering
        \includegraphics[width=0.9\columnwidth]{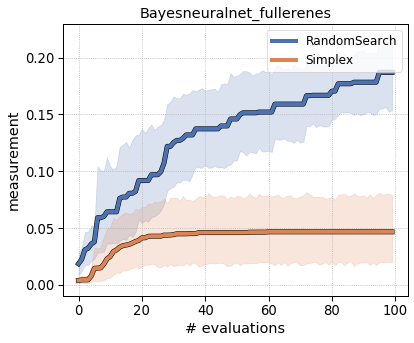}
        \caption{Plot generated with the \texttt{Plotter} module within \olympus illustrating the average best measurements collected during 5 independent runs with \random and \simplex over 100 iterations on the \olympuslabel{fullerenes} emulator.}
        \label{fig:olympus_plot}
    \end{figure}

    Results collected in several campaigns can be plotted automatically via a comprehensive plotting interface. Plots are generated from a \texttt{Database} object, like the one generated by \olympus when running a benchmark campaign. The following example illustrates the generation of a plot that illustrates the results of the executed benchmark. The generated plot is shown in Fig.~\ref{fig:olympus_plot}
    
\begin{minted}[mathescape, autogobble, numbersep=5pt, frame=lines, framesep=2mm, fontsize=\small]{python}
# plot collected results
from olympus import Olympus, Plotter
olymp = Olympus()
olymp.benchmark(num_iter=100,
    dataset='fullerenes',
    planners=['random','simplex'], 
    num_ind_runs=10)
>>> ...
Plotter().plot_from_db(olymp.database)
\end{minted}

    Further details on the capabilities and the usage of the \texttt{Plotter} module are reported in the online documentation.\cite{github:olympus}

    %=== MORE DETAILED STUDY WITH RANDOM SEARCH ===%

\section{A random search baseline} 

    The promise of data-driven strategies to identify desired parameter choices for experimental setups in closed-loop workflows is based on their capacity to condition design choices on feedback collected from previous experiments. The performance of a data-driven experiment planning strategies can, for example, be quantified via the number of experiments required to locate parameter values which yield the desired experimental outcomes. In this paragraph, we aim to provide a baseline for the performance of data-driven experiment planning strategies which can indicate the degree of difficulty that each constructed emulator poses to a planner.

    We construct the baseline by probing the performance of the random search strategy on each of the emulators. Random search as an experiment planning approach can be considered to be a naive strategy as it does not leverage any feedback collected in previous experiments. Parameter choices for future experiments are generated by drawing random samples from uniform distributions supported within the allowed ranges of each of the parameters. As such, the suggested parameter values are independent from one another and are not influenced by any past measurements. Data-driven strategies for experiment planning that do condition their design choices on previous feedback are therefore expected to outperform the random search baseline. The magnitude by which random search is outperformed can be used as a proxy to quantify the efficiency of the planner for each emulator.
    
    The provided baseline consists of 100 independent campaigns with 10,000 emulator evaluations per campaign for each of the emulators. Note that results from random baselines are not shipped with the software package and need to be downloaded separately. Random baselines are available on Github\cite{github:olympus} and can be downloaded from there or via the \olympus command line interface.
    
\begin{minted}[mathescape,autogobble,numbersep=5pt,frame=lines,framesep=2mm]{bash}
# download random baseline
>> olympus baseline --get
\end{minted}

    All parameters are generated with the random search planner. The results of these baseline calculations can be accessed through \olympus as follows
    
\begin{minted}[mathescape,autogobble,numbersep=5pt,frame=lines,framesep=2mm]{python}
# load the baseline
from olympus import Baseline
base      = Baseline()
summary   = base.get('snar', kind='summary')
campaigns = base.get('snar', kind='campaigns')
database  = base.get('snar', kind='db')
\end{minted}

    While the full traces of the random search baselines are available through \olympus, we suggest to compare the achieved feedback after a specified set of emulator evaluations. We propose to use [1, 3, 10, 30, 100, 300, 1000, 3000, 10000]. This choice is inspired by the fact that most experimental campaigns reported for autonomous experimentation platforms are limited to about 100 experiments. This set of evaluation numbers allows to estimate the performance of each planner in the regime of little data ($\sim$10 evaluations), medium data ($\sim$100 evaluations), abundant data ($\sim$1,000 evaluations) and asymptotic behavior ($\sim$10,000 evaluations). 
    
    \begin{figure}[!ht]
        \centering
        \includegraphics[width=1.0\columnwidth]{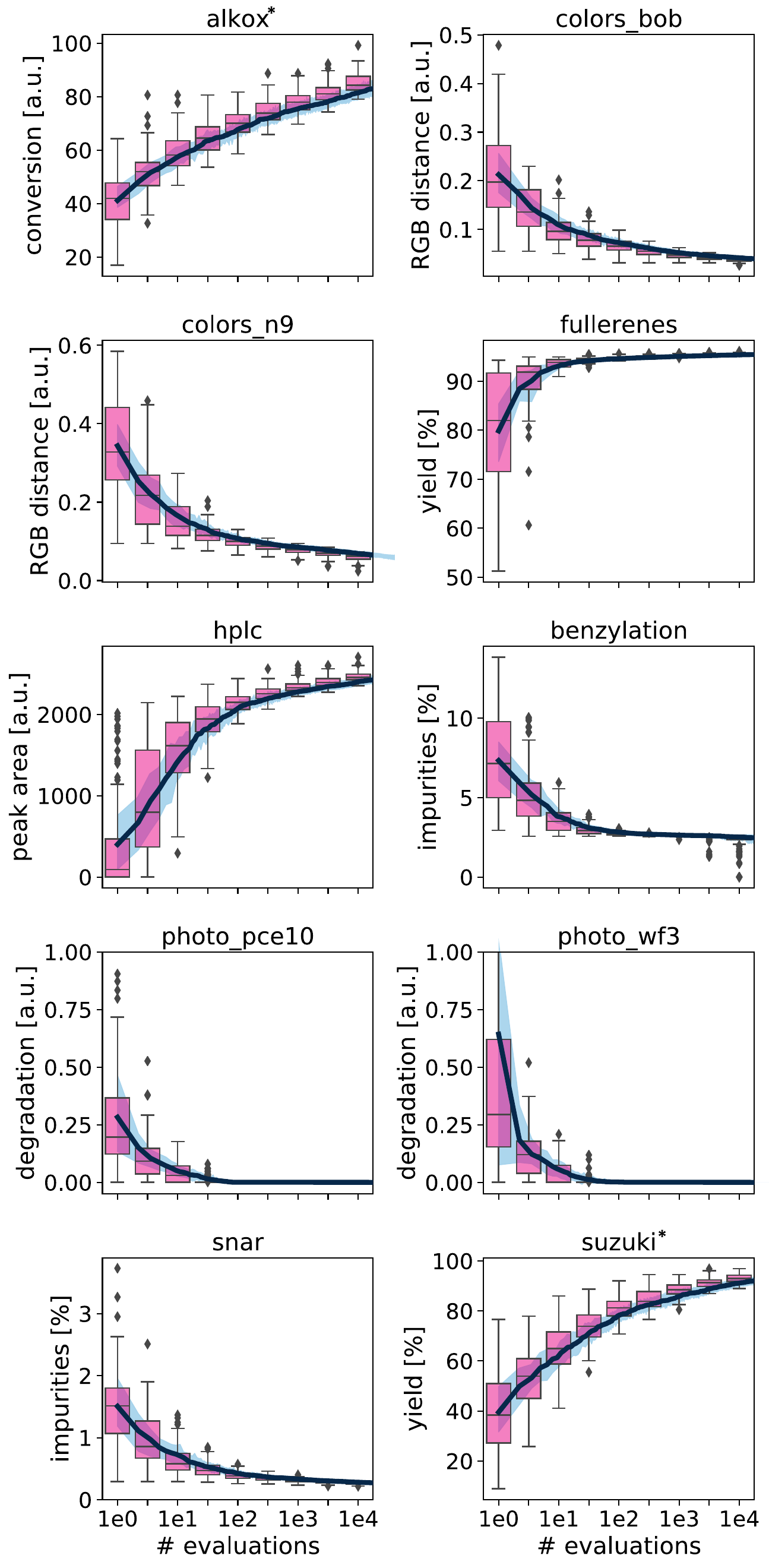}
        \caption{Performance of random search on all ten emulated surfaces in \olympus. We illustrate the best achieved property values up to the specified number of evaluations for all ten datasets. Solid blue lines show the average best values over 100 independent executions of the optimization starting from different random seeds. Box plots show the distribution of these 100 values at a few specific number of evaluations. Results obtained for emulators trained on datasets introduced in this work are marked with *.} 
        \label{fig:random_search_baseline}
    \end{figure}

    The results of random search against all core \olympus datasets are illustrated in Fig.~\ref{fig:random_search_baseline}. Based on these results, we can identify a subset of the emulated surfaces for which random search reaches near-optimal property values in a small number of evaluations. This subset includes \texttt{photobleaching\_pce10}, \texttt{photobleaching\_wf3}, \texttt{colormix\_bob}, \texttt{colormix\_n9}, \texttt{snar}, and \texttt{hplc\_n9}. Given that random search does not leverage any feedback from collected measurements for future decisions, these emulated surfaces might be considered to be the simpler cases for a more sophisticated experiment planner. The remaining surfaces, however, including \texttt{alkox}, \texttt{fullerenes}, \texttt{nbenzylation}, and \texttt{suzuki}, might pose a bigger challenge to experiment planners given that asymptotic property values are only achieved after a significant number of random evaluations or not even reached after 10,000 evaluations. Numerical values for the baseline are available through the \olympus package.\cite{github:olympus} 
    We hope that the results from this random search can serve as a baseline to compare the performance of different experiment planning strategies such as those already included in \olympus, but also new strategies developed by the community.

    %=== CONCLUSION ===%

\section{Conclusion}

    % summary of what we have done with olympus

    Standardized and challenging benchmarks are necessary to facilitate precise comparison between different approaches and allow scientific and technological advances to be quantified. Widely used benchmark sets like MNIST and CIFAR-10/100,\cite{Lecun:2010,cifar10} which are comprised of images of hand-written digits and various objects and animals, respectively, have allowed to measure constant advances in machine vision, providing clear feedback to the community on the most promising research directions. MoleculeNet, a collection of quantum mechanical, physical, biophysical and physiological molecular properties, provides a similar example in the field of chemistry and biophysics.\cite{Wu:2018} \olympus constitutes an orthogonal set of benchmarks, with a focus on optimization and experiment planning in chemistry and materials science, as opposed to prediction. It provides a framework with the potential to spark and streamline the development of powerful algorithms and data-driven approaches aimed at efficient experiment planning. To this end, \olympus also provides intuitive interfaces to a variety of experiment planning strategies to simplify their implementation, deployment, and testing in autonomous discovery workflows. With every user being able to supply their own datasets through our standardized interfaces, \olympus also encourages the free exchange of data across the community and promotes the establishment of standard, reproducible optimization challenges. In summary, \olympus provides a unified framework for the deployment and testing of experiment planning strategies. We thus invite the community to take advantage of \olympus in the implementation and testing of novel approaches to autonomous workflows, as well as to share experimental data that can prove valuable in moving this exciting new field forward.

	%=== ACKNOWLEDGEMENTS ===%
	\section*{Acknowledgments}

		The authors acknowledge generous support from Natural Resources Canada (NRCAN). F.H. acknowledges financial support from the Herchel Smith Graduate Fellowship and the Jacques-Emile Dubois Student Dissertation Fellowship. M.A. is supported by a Postdoctoral Fellowship of the Vector Institute. R.J.H. gratefully acknowledges the Natural Sciences and Engineering Research Council of Canada (NSERC) for provision of the Postgraduate Scholarships-Doctoral Program (PGSD3-534584-2019). 
This work relates to Department of Navy award (N00014-19-1-2134) issued by the Office of Naval Research. This work was supported by the Defense Advanced Research Projects Agency under the Accelerated Molecular Discovery Program under Cooperative Agreement No. HR00111920027 dated August 1, 2019. The content of the information presented in this work does not necessarily reflect the position or the policy of the Government. A.A.G. would like to thank Dr.~Anders Fr{\o}seth for his support. All computations reported in this paper were completed on the computing clusters of the Vector Institute and the Odyssey cluster supported by the FAS Division of Science, Research Computing Group at Harvard University. Resources used in preparing this research were provided, in part, by the Province of Ontario, the Government of Canada through CIFAR, and companies sponsoring the Vector Institute.

	%=== BIBLIOGRAPHY ===%

%	\phantomsection\addcontentsline{toc}{section}{\refname}\putbib[main]

    \bibliography{main}

\clearpage
\newpage

%\begin{bibunit}[unsrt]

	\onecolumngrid
	\setcounter{subsection}{0}

\section{Supplementary information}
    
    \subsection{List of datasets}
    \label{supp_sec:datasets}
        
        In this section we provide a brief summary of each dataset available in \olympus, along with the parameters, objectives and optimization goal.  
    
        % =============
        % Alkox
        % =============
        \subsubsection{Alkoxylation}
        
            This dataset contains 104 measurements on biocatalytic oxidation of benzyl alcohol by a copper radical oxidase (AlkOx). The effects of enzyme loading, cocatalyst loading, and pH balance on both initial rate and total conversion were assayed. Stock solution were prepared daily and stored over crushed ice. Additional dilutions were done as required using sodium phosphate buffer and immediately discarded after use. The assays were initiated by the addition of \emph{Cgr}AlcOx and H$_2$O$_2$ to a well-mixed HPLC vial containing all other reaction components. The initial rate was obtained by fitting the concentration of aldehyde to a linear function and reporting the slope. Conversion was calculated by fitting the percent conversion of aldehyde to a linear function and reporting the value at twenty minutes. 
            
            \begin{figure}[!ht]
                \centering
                \includegraphics[width=0.5\textwidth]{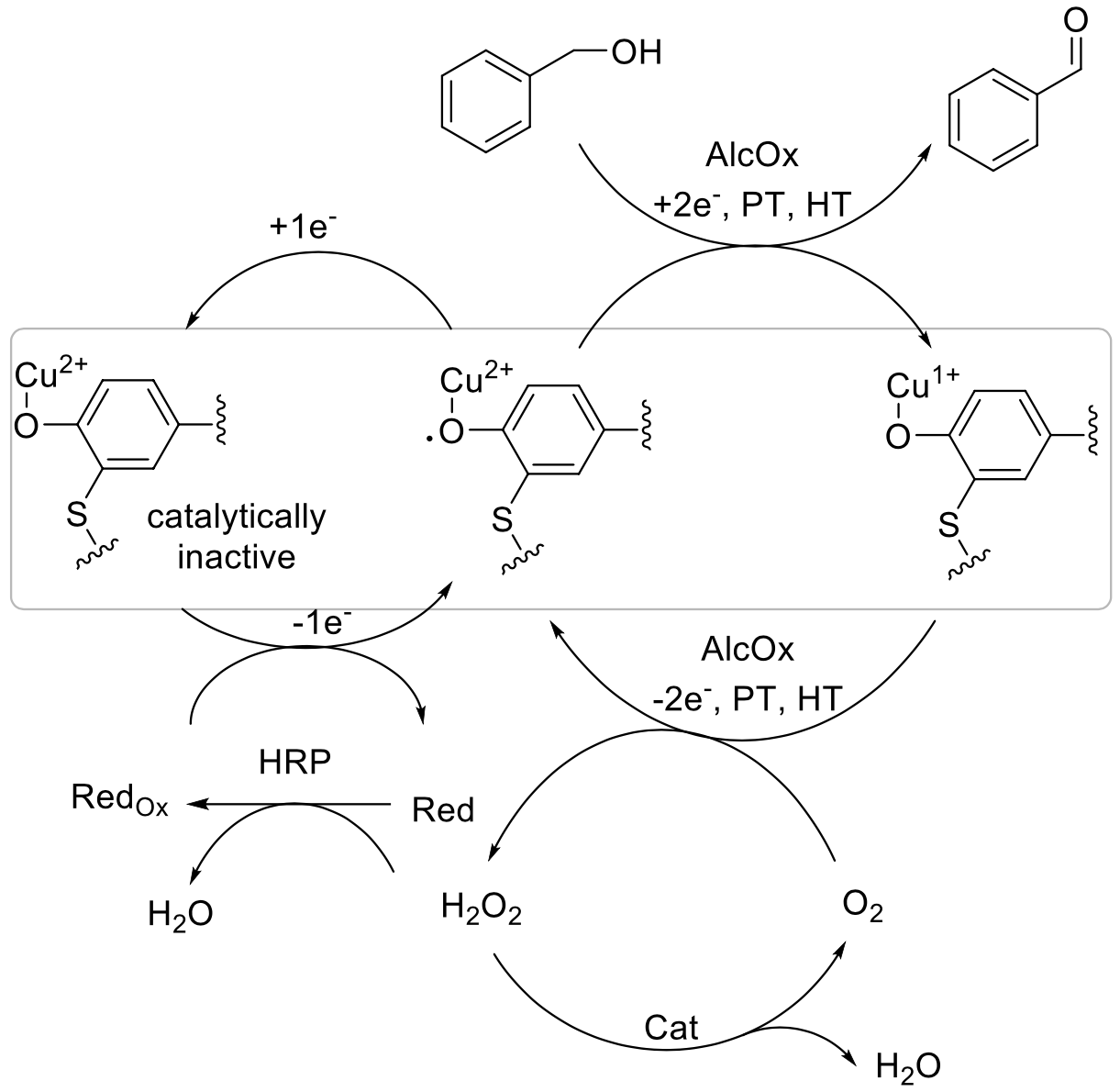}
                \caption{Scheme of the AlcOx assay.}
                \label{supp_fig:alkox_scheme}
            \end{figure}
                
            \begin{table}[!ht]
                \centering
                \caption{Parameter space of the alkoxylation dataset}
                \label{supp_tab:alkoxylation_params}
                \begin{tabular}{c c c c c} 
                    \cline{1-5}
                    \textbf{Parameter} & \textbf{Kind} & \textbf{Range} & \textbf{Description} & \textbf{Objective} \\ %[0.5ex] 
                    \cline{1-5}
                    Catalase                & continuous & [0.05, 1.0] & concentration [$\mu$M] & \multirow{4}{*}{conversion $\uparrow$}\\ 
                    Horseradish peroxidase  & continuous & [0.5, 10.0] & concentration [$\mu$M] \\ 
                    Alcohol oxidase         & continuous & [2.0, 8.0] & concentration [nM] \\ 
                    pH & continuous & [6.0, 8.0] & -log(H$^+$) [ml/min] \\ 
                    \cline{1-5}
                \end{tabular}
                \label{supp_tab:alkox_params}
            \end{table}

        % =============
        % Color Bob
        % =============
        \subsubsection{Color mixing BOB}
        
        This dataset consists of colors prepared by mixing varying amounts of 5 colored dyes (red, orange, yellow, blue, green). The parameters represent fractions of each dye used in a mixture. The target is the normalized green-like RGB value [0.16, 0.56, 0.28]. This dataset consists of 241 measurements performed using the Bayesian Optimized Bartender (BOB). \cite{Roch:2020}
        
        \begin{table}[!ht]
        \centering
                \caption{Parameter space of the color BOB dataset}
                \label{supp_tab:color_bob_params}
         \begin{tabular}{ccccc} 
         \cline{1-5}
         \textbf{Parameter} & \textbf{Kind} & \textbf{Range} & \textbf{Description} & \textbf{Objective} \\ %[0.5ex] 
         \cline{1-5}
         red & continuous & [0, 1] & amount of red & \multirow{5}{*}{distance to green $\downarrow$} \\ 
         orange & continuous & [0, 1] & amount of orange \\
         yellow & continuous & [0, 1] & amount of yellow \\
         blue & continuous & [0, 1] & amount of blue \\
         green & continuous & [0, 1] & amount of green \\
         \cline{1-5}
         \end{tabular}
        \end{table}
        
        % =============
        % Color N9
        % =============
        \subsubsection{Color mixing N9}
        
        This dataset consists of colors prepared by mixing 3 types of dyes (red, green and blue). The target is the normalized green-like RGB value [0.16, 0.56, 0.28]. This dataset consists of 102 measurements performed with an N9 robotic arm from North Robotics.
        
        \begin{table}[!ht]
        \centering
                \caption{Parameter space of the color N9 dataset}
                \label{supp_tab:color_n9_params}
         \begin{tabular}{ccccc} 
         \cline{1-5}
         \textbf{Parameter} & \textbf{Kind} & \textbf{Range} & \textbf{Description} & \textbf{Objective} \\ %[0.5ex] 
         \cline{1-5}
         red & continuous & [0, 1] & amount of red & \multirow{3}{*}{distance to green $\downarrow$}\\ 
         green & continuous & [0, 1] & amount of green \\
         blue & continuous & [0, 1] & amount of blue \\
         \cline{1-5}
         \end{tabular}
        \end{table}
        
        % =============
        % Fullerenes
        % =============
        \subsubsection{Buckminsterfullerene adducts}
        
\begin{figure}[!ht]
            \centering
            \includegraphics[width=0.8\textwidth]{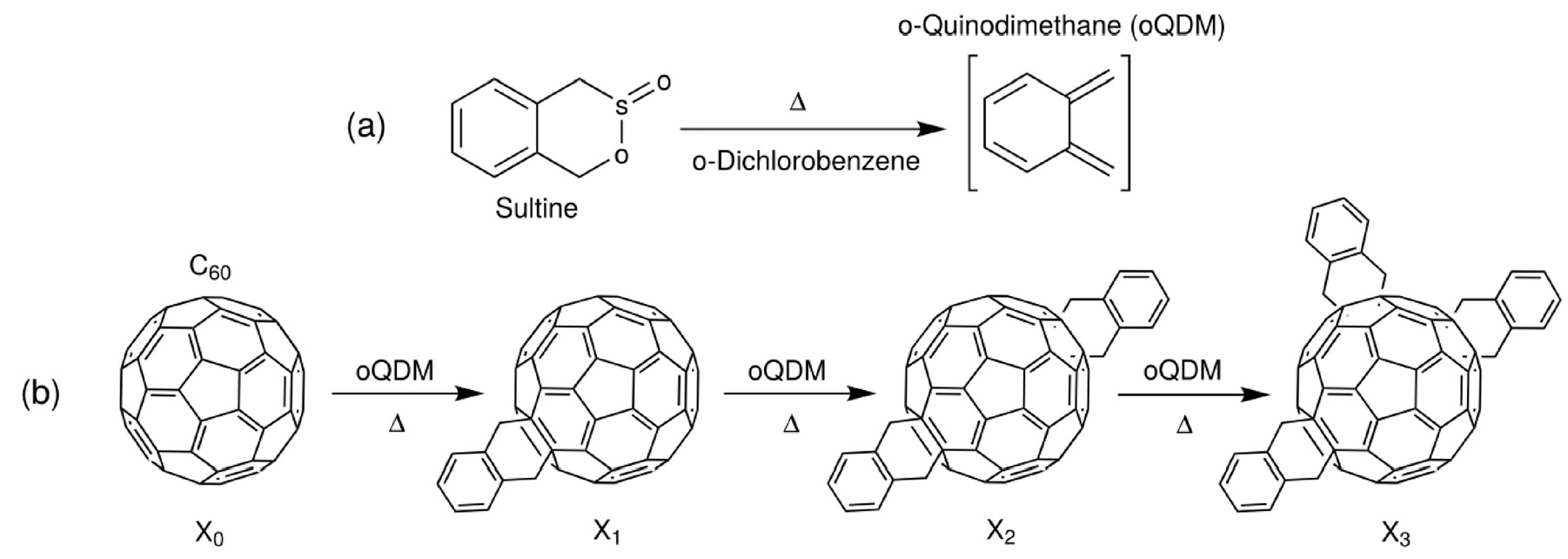}
            \caption{Synthesis of o-xylenyl C\textsubscript{60} adducts of varying order via (a) in situ conversion of sultine (1,4-dihydro-2,3-benzoxathiin 3-oxide) to o-quinodimethane (oQDM)  followed by (b) successive attachments of oQDM to C\textsubscript{60} (X0) by Diels–Alder cycloadditions. Reproduced from Walker et al.\cite{Walker:2017:fullerenes} under a CC BY 3.0 license.}
            \label{supp_fig:fullerenes_scheme}
        \end{figure}

        This dataset is based on the reported production of o-xylenyl adducts of Buckminsterfullerenes (Fig. \ref{supp_fig:fullerenes_scheme}).~\cite{Walker:2017:fullerenes} Three process conditions can be varied to maximize the mole fraction of the desired products X\textsubscript{1} and X\textsubscript{2}. The conditions are temperature, reaction time and the ratio of sultine to C\textsubscript{60}. Experiments were based on a fully factorial design with three factors and six levels, totalling 246 samples.
        
        \begin{table}[!ht]
        \centering
                \caption{Parameter space of the Buckminsterfullerene dataset}
                \label{supp_tab:buckminsterfullerene_params}
         \begin{tabular}{ccccc} 
         \cline{1-5}
         \textbf{Parameter} & \textbf{Kind} & \textbf{Range} & \textbf{Description} & \textbf{Objective} \\ %[0.5ex] 
         \cline{1-5}
         reaction time & continuous & [3, 31] & reaction time in flow reactor [min] & \multirow{3}{*}{mole fraction of X\textsubscript{1} + X\textsubscript{2} $\uparrow$} \\ 
         sultine conc & continuous & [1.5, 6.0] & relative concentration of sultine to C\textsubscript{60} \\
         temperature & continuous & [100, 150] & temperature of the reaction [deg Celsius] \\
         \cline{1-5}
         \end{tabular}
        \end{table}

        % =============
        % HPLC N9
        % =============
        \subsubsection{HPLC}
        
        This dataset reports the peak response of an automated high-performance liquid chromatography (HPLC) system for varying process parameters.~\cite{Roch:2020} The dataset includes 1,386 samples with six parameters and one objective.
        
        \begin{table}[!ht]
        \centering
                \caption{Parameter space of the HPLC dataset}
                \label{supp_tab:hplc_params}
         \begin{tabular}{c c c c c} 
         \cline{1-5}
         \textbf{Parameter} & \textbf{Kind} & \textbf{Range} & \textbf{Description} & \textbf{Objective} \\ 
         \cline{1-5}
         sample loop & continuous & [0.00, 0.08] & volume of the sample loop [ml] & \multirow{6}{*}{peak area $\uparrow$}\\ 
         additional volume & continuous & [0.00, 0.06] & volume required to draw sample [ml] \\ 
         tubing volume & continuous & [0.1, 0.9] & volume required to drive sample [ml] \\ 
         sample flow & continuous & [0.5, 2.5] & draw rate of sample pump [ml/min] \\ 
         push speed  & continuous & [80, 150] & draw rate of push pump [Hz] \\ 
         wait time & continuous & [1, 10] & wait time [s] \\ 
          \cline{1-5}
         \end{tabular}
        \end{table}
    
        % =============
        % N-Benzyl
        % =============
        \subsubsection{N-benzylation}
\begin{figure}[!ht]
                \centering
                \includegraphics[width=0.7\textwidth]{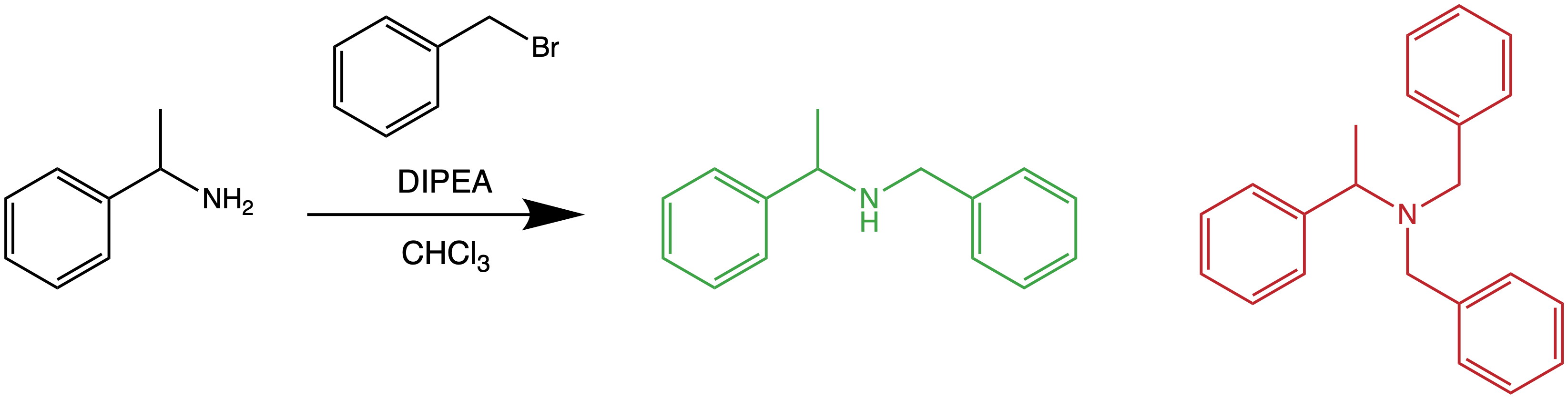}
                \caption{Scheme of the N-benzylation reaction. The desired secondary amine is shown in green, while the undesired tertiary amine is shown in red.}
                \label{supp_fig:nbenzylation_scheme}
        \end{figure}

        This dataset reports the yield of undesired product (impurity) in an N-benzylation reaction.~\cite{Schweidtmann:2018} The undesired product is the tertiary amine shown in Fig. \ref{supp_fig:nbenzylation_scheme}. This dataset includes 73 samples with four parameters and one objective.
        
        \begin{table}[!ht]
                \centering
                \caption{Parameter space of the N-benzylation dataset}
                \label{supp_tab:nbenzylation_params}
                \begin{tabular}{ccccc} 
                \cline{1-5}
                \textbf{Parameter} & \textbf{Kind} & \textbf{Range} & \textbf{Description} & \textbf{Objective} \\ %[0.5ex] 
                \cline{1-5}
                flow rate & continuous & [0.2, 0.4] & flow rate [mL/min] & \multirow{4}{*}{impurity yield $\downarrow$}\\ 
                ratio & continuous & [1, 5] & benzyl bromide equivalents w.r.t. the methylbenzylamine reagent \\ 
                solvent & continuous & [0.5, 1.0] & solvent (CHCl\textsubscript{3}) equivalents w.r.t. stock solution of reagent \\ 
                temperature & continuous & [110, 150] & reaction temperature [\textdegree{}C] \\ 
                \cline{1-5}
                \end{tabular}
            \end{table}

        % =============
        % Photobleaching PCE10
        % =============   
        \subsubsection{Photobleaching PCE10}
        This dataset reports the degradation of polymer blends for organic solar cells under the exposure to light. Individual data points encode the ratios of individual polymers in one blend, along with the measured photodegradation of this blend.~\cite{Langner:2020} The dataset includes 1040 samples with four parameters and one objective. 
        
        \begin{table}[!ht]
        \centering
                \caption{Parameter space of the photobleaching PCE10 dataset}
                \label{supp_tab:photo_pce10_params}
         \begin{tabular}{ccccc} 
         \cline{1-5}
         \textbf{Parameter} & \textbf{Kind} & \textbf{Range} & \textbf{Description} & \textbf{Objective} \\ %[0.5ex] 
         \cline{1-5}
         mat 1 & continuous & [0, 1] & amount of PCE10 & \multirow{4}{*}{photo-degradation $\downarrow$} \\ 
         mat 2 & continuous & [0, 1] & amount of P3HT \\
         mat 3 & continuous & [0, 1] & amount of PCBM \\
         mat 4 & continuous & [0, 1] & amount of olDTBR \\
         \cline{1-5}
         \end{tabular}
        \end{table}
        
         % =============
        % Photobleaching WF3
        % =============   
        \subsubsection{Photobleaching WF3}
    
        This dataset reports the degradation of polymer blends for organic solar cells under the exposure to light. Individual data points encode the ratios of individual polymers in one blend, along with the measured photodegradation of this blend.~\cite{Langner:2020} The dataset includes 1040 samples with four parameters and one objective.
        
        \begin{table}[!ht]
        \centering
                \caption{Parameter space of the photobleaching WF3 dataset}
                \label{supp_tab:photo_wf3_params}
         \begin{tabular}{ccccc} 
         \cline{1-5}
         \textbf{Parameter} & \textbf{Kind} & \textbf{Range} & \textbf{Description} & \textbf{Objective} \\ %[0.5ex] 
         \cline{1-5}
         mat 1 & continuous & [0, 1] & amount of WF3 & \multirow{4}{*}{photo-degradation $\downarrow$} \\ 
         mat 2 & continuous & [0, 1] & amount of P3HT \\
         mat 3 & continuous & [0, 1] & amount of PCBM \\
         mat 4 & continuous & [0, 1] & amount of olDTBR \\
         \cline{1-5}
         \end{tabular}
        \end{table}

        % =============
        % SNAR
        % =============
        \subsubsection{SnAr reaction}

\begin{figure}[!ht]
                \centering
                \includegraphics[width=0.7\textwidth]{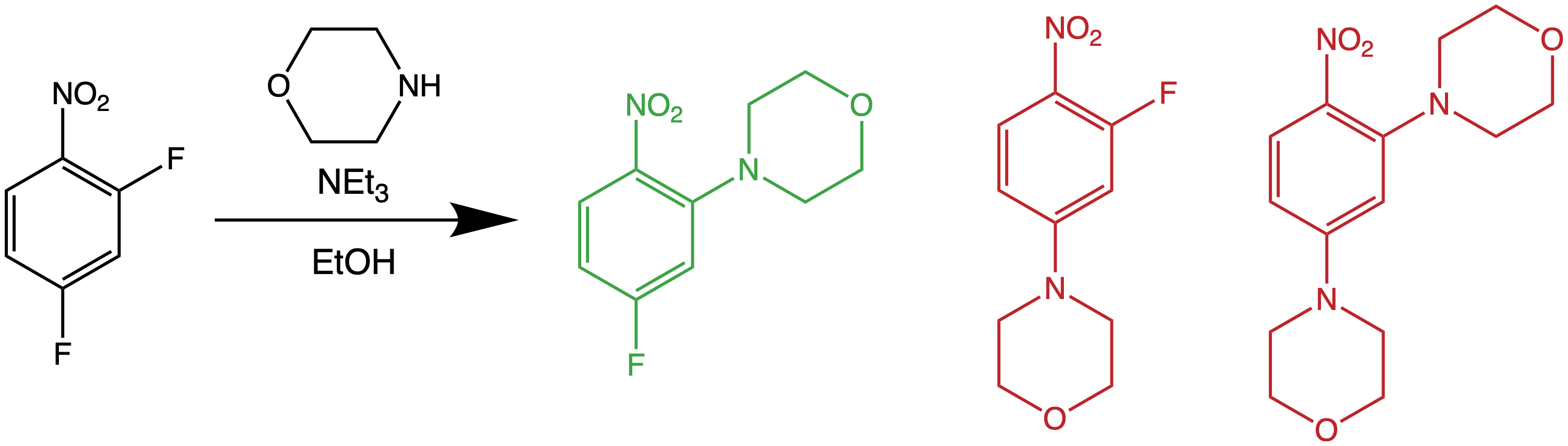}
                \caption{Scheme of the S\textsubscript{N}Ar reaction. The desired ortho product is shown in green, while the undesired para and bis adduct products are shown in red.}
                \label{supp_fig:snar_scheme}
        \end{figure}

        This dataset reports the environmental factor (E-factor) for the nucleophilic aromatic substitution (S\textsubscript{N}Ar) reaction shown in Fig. \ref{supp_fig:snar_scheme}.~\cite{Schweidtmann:2018} The E-factor is defined as the ratio of the mass of waste to the mass of product. The dataset includes 66 samples with four parameters and one objective.
        
        \begin{table}[!ht]
        \centering
                \caption{Parameter space of the SNAr dataset}
                \label{supp_tab:snar_params}
         \begin{tabular}{ccccc} 
         \cline{1-5}
         \textbf{Parameter} & \textbf{Kind} & \textbf{Range} & \textbf{Description} & \textbf{Objective}\\ %[0.5ex] 
         \cline{1-5}
         residence time & continuous & [0.5, 2.0] & residence time for flow apparatus [min] & \multirow{4}{*}{E-factor $\downarrow$} \\ 
         morpholine equiv & continuous & [1.0, 5.0] & morpholine equivalents \\
         concentration & continuous & [0.1, 0.5] & concentration of reagents [M] \\
         temperature & continuous & [60.0, 140.0] & temperature of the reactor [Celsius] \\
         \cline{1-5}
         \end{tabular}
        \end{table}
        
        % =============
        % Suzuki
        % =============
        \subsubsection{Suzuki reaction}
\begin{figure}[!ht]
                \centering
                \includegraphics[width=0.5\textwidth]{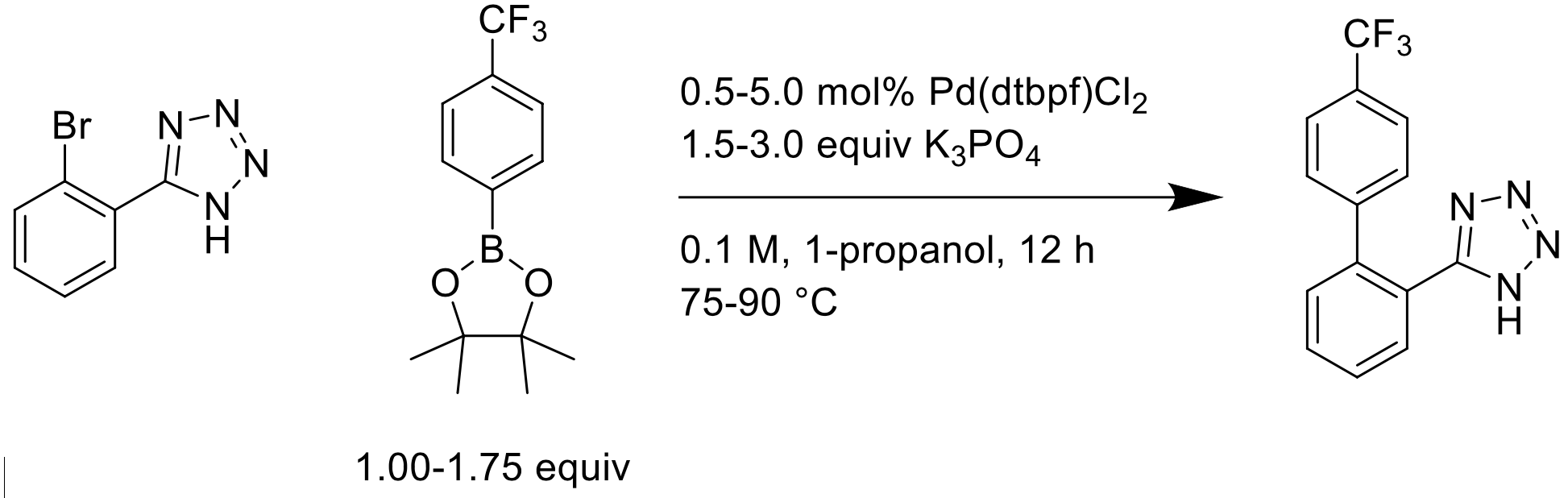}
                \caption{Scheme of the Suzuki reaction}
                \label{supp_fig:suzuki_scheme}
            \end{figure}

            High-throughput reactions were carried out on the palladium-catalyzed Suzuki cross-coupling between 2-bromophenyltetrazole and an electron-deficient aryl boronate (see Fig.~\ref{supp_fig:suzuki_scheme}). Cross-couplings of aryl halide electrophiles bearing non-protected ortho tetrazole substituents are typically carried out under harsh conditions due to the metal-chelating nature of the tetrazole moiety, but we have found that the use of electron-rich bidentate phosphines, such as dtbpf, in alcohol solvents, facilitate milder reaction conditions. 
            
            A wide range of continuous factors were explored in the microscale optimization run, including reaction temperature, Pd(dtbpf)Cl$_2$ loading, and equivalents of base. This resulted in product yields spanning from 2 to 97 mol \%, generating a robust data set for modeling. Parameters and ranges are summarized in Tab.~\ref{supp_tab:suzuki_params}
            
            \begin{table}[!ht]
                \centering
                \caption{Parameter space of the Suzuki dataset}
                \label{supp_tab:suzuki_params}
                \begin{tabular}{ccccc} 
                \cline{1-5}
                \textbf{Parameter} & \textbf{Kind} & \textbf{Range} & \textbf{Description} & \textbf{Objective} \\ %[0.5ex] 
                \cline{1-5}
                temperature & continuous & [75, 90] & temperature of the reaction [\textdegree{}C] & \multirow{4}{*}{yield $\uparrow$} \\ 
                Pd mol & continuous & [0.5, 5.0] & loading of Pd catalyst [mol \%] \\ 
                ArBpin & continuous & [1.0, 1.8] & equivalents of pinacolyl boronate ester coupling partner \\ 
                K\textsubscript{3}PO\textsubscript{4} & continuous & [1.5, 3] & equivalents of tripotassium phosphate \\ 
                \cline{1-5}
                \end{tabular}
            \end{table}

    \subsection{Description of emulators}
    \label{supp_sec:emulators}
    % accuracy plots moved to last subsection
    
\begin{figure}[!ht]
        \centering
        \includegraphics[width=\textwidth]{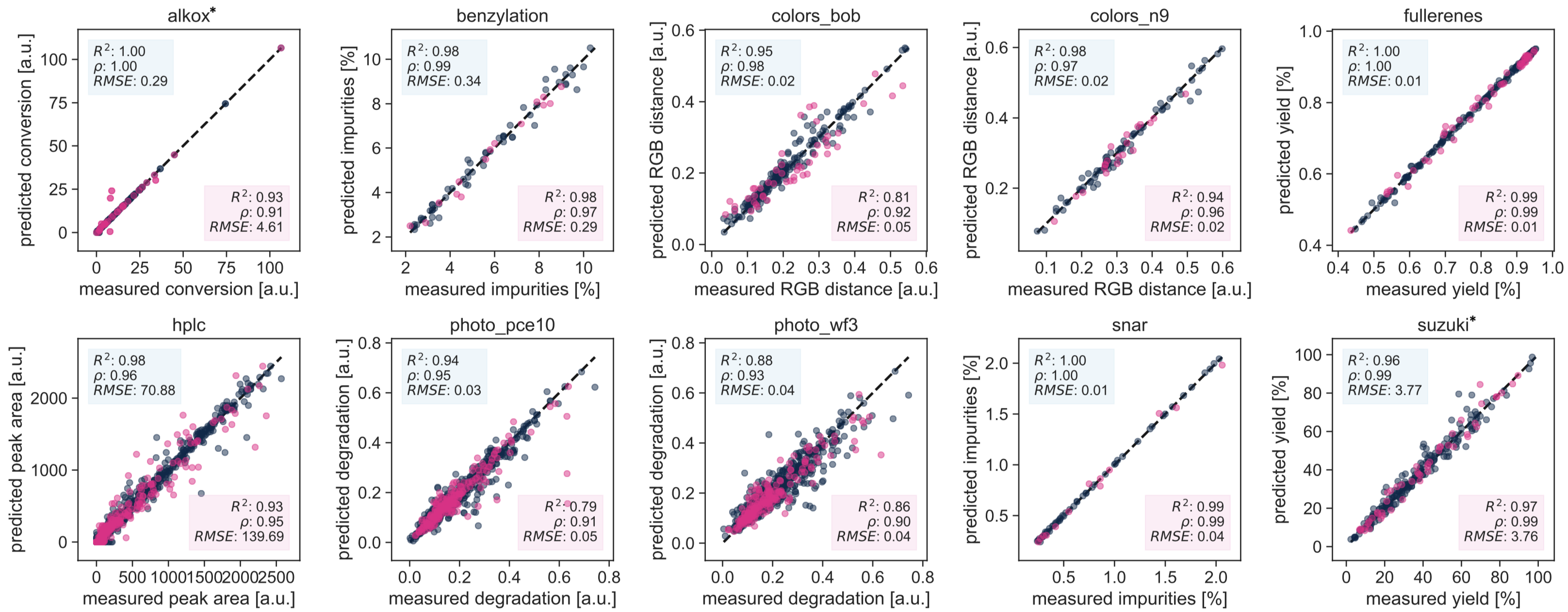}
        \caption{Parity plots with experimental versus predicted target values for all emulators based on Neural Network models. Performance on the training set (80\% of data; blue markers) is shown in the top-left corner of each plot and on blue background; performance on the test set (20\% of data; pink markers) is shown in the bottom-right corner of each plot and on pink background. $R^2$ is the coefficient of determination, $\rho$ is the Spearman's rank correlation coefficient, and \textit{RMSE} is the root-mean-square error. Emulators trained on datasets introduced in this study are indicated with *.}
        \label{si_fig:nn_emulators}
    \end{figure}

    The datasets in \olympus can be emulated using probabilistic Bayesian neural network (BNN) models or deterministic neural network (NN) models. Fig. \ref{fig:bnn_emulators} and Fig. \ref{si_fig:nn_emulators} show the correlation between predicted and measured target data points for BNN and NN models, respectively. All emulators display a Spearman's rank coefficient above $0.90$, for both training and test sets. Train/test splits were performed at random, but using a fixed random seed for reproducibility; 80\% of the data was used for training and 20\% for testing. Hyperparameter optimization was performed manually using 5-fold cross validation. The details of all hyperparameters used in these models are stored in the respective \texttt{Emulator} objects that can be loaded from \olympus.
    
    %\begin{figure}[h!]
    %    \centering
    %    \includegraphics[width=\textwidth]{figures/emulators_bnn_wlabels_si.pdf}
    %    \caption{Parity plots with experimental versus predicted target values for all emulators based on Bayesian Neural Network models. Performance on the training set (80\% of data; blue markers) is shown in the top-left corner of each plot and on blue background; performance on the test set (20\% of data; pink markers) is shown in the bottom-right corner of each plot and on pink background. $R^2$ is the coefficient of determination, $\rho$ is the Spearman's rank correlation coefficient, and \textit{RMSE} is the root-mean-square error.}
    %    \label{si_fig:bnn_emulators}
    %\end{figure}
    
    \subsection{Description of surfaces}
    \label{supp_sec:surfaces}
    
\begin{figure}[!ht]
        \begin{center}
            \includegraphics[width=\textwidth]{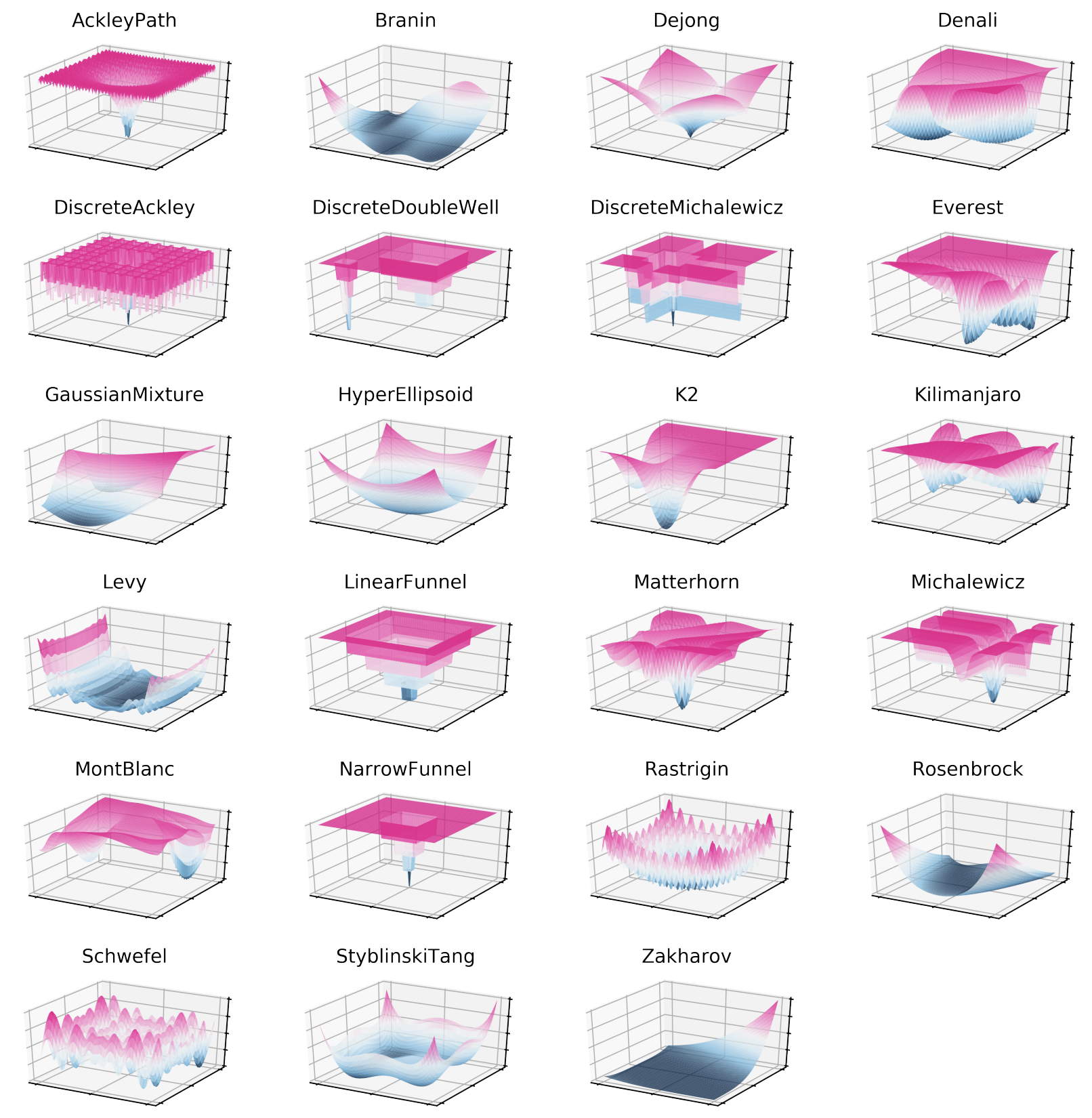}
            \end{center}
            \caption{Analytical surfaces available in \olympus.}%
            \label{si_fig:surfaces}%
    \end{figure}

    Fig. \ref{si_fig:surfaces} illustrates the analytical benchmark surfaces available in \olympus. With the exception of \texttt{Branin}, which is restricted to a two-dimensional input space, all other surfaces may be defined in any dimension. Note that all surfaces operate on the unit hypercube, $\mathcal{X} \in [0,1]^d$. \olympus internally scales the inputs to be in agreement with the untransformed domains typically employed for these analytical functions (Table \ref{si_tab:surfaces}).
    
    \begin{table}[!ht]
        \centering
         \caption{Details of the continuous analytical surfaces available in \olympus.}
         \begin{tabular}{p{2.8cm} p{2.4cm} p{2cm} p{10cm}}
         \cline{1-4}
         Surface & Domain & Default parameters & {Formula\newline$f(\mathbf{x})=$} \\ %[0.5ex] 
         \cline{1-4}
         \texttt{AckleyPath} & $\mathbf{x} \in [-32,32]^d$ & $d=2$ & $-20 \cdot \exp \Big[ -0.2 \big(\frac{1}{d} \sum\limits_{i=1}^{d} {x_i^2}\big)^{\frac{1}{2}} \Big] - \exp \Big( \sum\limits_{i=1}^{d} \cos{(2\pi x_i)} \Big) + 20 + \exp(1)$ \\ 
         \texttt{Branin}  &  {$x_1 \in [-5,10]$ \newline $ x_2 \in [0,15]$} & n.a. & {$(x_2 - bx_1 + cx_1 - 6)^2 +10(1-t)\cos(x_1) + 10$ \newline where $b=5.1/4\pi^2,\ c=5/\pi,\ t=1/8\pi$} \\
         \texttt{Dejong}  &  $\mathbf{x} \in [-5,5]^d$ & $d=2$ & $\sum\limits_{i=1}^{d} |x_i|^{\frac{1}{2}}$ \\
         \texttt{HyperEllipsoid}  &  $\mathbf{x} \in [-5, 5]^d$ & $d=2$ & $\sum\limits_{i=1}^{d} i x_i^2$ \\
         \texttt{Levy}  &  $\mathbf{x} \in [-10,10]^d$ &  $d=2$ & {$\sin^2(\pi w_1)+\sum\limits_{i=1}^{d-1}(w_i-1)^2[1 + 10\sin^2(\pi w_i +1)]+(w_d-1)[1+\sin^2(2\pi w_d)]$ \newline where $w_i = 1 + (x_i-1)/4$} \\
         \texttt{Michalewicz}  &  $\mathbf{x} \in [0,\pi]^d$ & {$d=2$\newline$m=10$} & $-\sum\limits_{i=1}^{d}\sin(x_i) \sin^{2m}(ix_i^2 \pi^{-1})$ \\
         \texttt{Rastrigin}  &  $\mathbf{x} \in [-5,5]^d$ & $d=2$ & $10d + \sum\limits_{i=1}^{d}[x_i^2 - 10\cos(2\pi x_i)]$ \\
         \texttt{Rosenbrock}  &  $\mathbf{x} \in [-2,2]^d$ & $d=2$ & $\sum\limits_{i=1}^{d-1}[100(x_{i+1}-x_i^2)^2 + (x_i -1)^2]$ \\
         \texttt{Schwefel}  &  $\mathbf{x} \in [-500,500]^d$ & $d=2$ & $-\sum\limits_{i=1}^{d}x_i\sin(|x_i|^{\frac{1}{2}})$ \\
         \texttt{StyblinskiTang}  &  $\mathbf{x} \in [-5,5]^d$ & $d=2$ & $0.5\sum\limits_{i=1}^{d}(x_i^4 - 16x_i^2 +5x_i)$ \\
         \texttt{Zakharov}  &  $\mathbf{x} \in [-5,10]^d$ & $d=2$ & $\sum\limits_{i=1}^{d}x_i^2 + \Big(\sum\limits_{i=1}^{d} 0.5ix_i \Big)^2 + \Big(\sum\limits_{i=1}^{d} 0.5ix_i \Big)^4 $ \\
         \cline{1-4}
         \end{tabular}
         \label{si_tab:surfaces}%
        \end{table}

%    \subsection{Description of random search baseline}
%    \label{supp_sec:baseline}

%	\putbib[main]
%\end{bibunit}

\end{document}